%% file: arxiv.tex
\documentclass[10pt,twocolumn,letterpaper]{article}

\usepackage[pagenumbers]{cvpr} 
\usepackage{graphicx}
\usepackage{amsmath}
\usepackage{amssymb}
\usepackage{booktabs}
\usepackage{array}
\usepackage{graphicx, amsmath, amssymb, caption, subcaption, multirow, overpic, textpos}
\usepackage[table]{xcolor}
\usepackage{colortbl}
\usepackage[linesnumbered,ruled,vlined]{algorithm2e}
\SetKwInOut{Parameter}{Parameters}
\usepackage{listings}
\usepackage{enumitem}

\makeatletter
\@namedef{ver@everyshi.sty}{}
\makeatother
\usepackage{tikz}

\newcommand*\circled[1]{\tikz[baseline=(char.base)]{
            \node[shape=circle,draw,inner sep=.5pt] (char) {#1};}}
\usepackage[pagebackref]{hyperref}
\hypersetup{breaklinks,colorlinks}
\definecolor{darkred}{rgb}{0.8,0.02,0.02}

\usepackage[capitalize]{cleveref}
\crefname{section}{Sec.}{Secs.}
\Crefname{section}{Section}{Sections}
\Crefname{table}{Table}{Tables}
\crefname{table}{Tab.}{Tabs.}

\def\cvprPaperID{5220} 
\def\confName{CVPR}
\def\confYear{2023}

\renewcommand{\paragraph}[1]{\vspace{0.6mm}\noindent\textbf{#1}}
\newcolumntype{x}[1]{>{\centering\arraybackslash}p{#1pt}}
\newcolumntype{y}[1]{>{\raggedright\arraybackslash}p{#1pt}}
\newcolumntype{z}[1]{>{\raggedleft\arraybackslash}p{#1pt}}
\newcommand{\app}{\raise.17ex\hbox{$\scriptstyle\sim$}}

\definecolor{deemph}{gray}{0.6}

\definecolor{baselinecolor}{gray}{.9}

\newcommand{\figref}[1]{Fig.~\ref{#1}}
\newcommand{\tabref}[1]{Table~\ref{#1}}
\newcommand{\equref}[1]{Eq.~(\ref{#1})}
\newcommand{\secref}[1]{Section~\ref{#1}}
\def\loss{\mathcal{L}\xspace}
\newcommand{\drule}{\specialrule{0.2pt}{1pt}{1pt}%
            \specialrule{0.2pt}{0pt}{\belowrulesep}%
            }

\begin{document}

\title{EcoTTA: Memory-Efficient Continual Test-time Adaptation \\via Self-distilled Regularization}
\newcommand*{\affmark}[1][*]{\textsuperscript{#1}}
\newcommand*{\email}[1]{\texttt{#1}}

\author{
{Junha Song}\affmark[1,2]\thanks{Work done during an internship at Qualcomm AI Research.} \,,\,\;
{Jungsoo Lee}\affmark[1],\,\;
{In So Kweon}\affmark[2],\,\;
{Sungha Choi}\affmark[1]\thanks{Corresponding author.\;\quad$^\ddag$ Qualcomm AI Research is an initiative of Qualcomm Technologies, Inc.}
\\
\affmark[1]Qualcomm AI Research$^\ddag$,\;\;
\affmark[2]KAIST\\
}
\maketitle

\begin{abstract}
\input{section/abstract}
\end{abstract}

\section{Introduction}
\label{sec:intro}
\input{section/intro}

\section{Related Work}
\label{sec:related}
\input{section/related}

\section{Approach}
\label{sec:approach}
\input{section/approach}

\section{Classification Experiments}
\label{sec:experiment1}
\input{section/experiment1}

\vspace{-.4em}
\section{Segmentation Experiments}
\label{sec:experiment2}
\vspace{-.4em}
\input{section/experiment2}

\vspace{-.4em}
\section{Conclusion}
\label{sec:conclusion}
\vspace{-.4em}
\input{section/conclusion}

\vspace{.3em}
\paragraph{Acknowledgments.} We would like to thank Kyuwoong Hwang, Simyung Chang, and Byeonggeun Kim for their valuable feedback. We are also grateful for the helpful discussions from Qualcomm AI Research teams. 

\newpage
{\small
\bibliographystyle{ieee_fullname}
\bibliography{egbib}
}

\clearpage
\newpage
\appendix
\input{section/appendix}

\end{document}

%% file: section/abstract.tex
This paper presents a simple yet effective approach that improves continual test-time adaptation (TTA) in a memory-efficient manner. 
TTA may primarily be conducted on edge devices with limited memory, so reducing memory is crucial but has been overlooked in previous TTA studies. In addition, long-term adaptation often leads to catastrophic forgetting and error accumulation, which hinders applying TTA in real-world deployments. Our approach consists of two components to address these issues. 
First, we present lightweight meta networks that can adapt the frozen original networks to the target domain. This novel architecture minimizes memory consumption by decreasing the size of intermediate activations required for backpropagation.
Second, our novel self-distilled regularization controls the output of the meta networks not to deviate significantly from the output of the frozen original networks, thereby preserving well-trained knowledge from the source domain.
Without additional memory, this regularization prevents error accumulation and catastrophic forgetting, resulting in stable performance even in long-term test-time adaptation. 
We demonstrate that our simple yet effective strategy outperforms other state-of-the-art methods on various benchmarks for image classification and semantic segmentation tasks. Notably, our proposed method with ResNet-50 and WideResNet-40 takes 86\% and 80\% less memory than the recent state-of-the-art method, CoTTA.
\vspace*{-0.1cm}

%% file: section/intro.tex
\begin{figure}[t]\centering
\includegraphics[width=1\linewidth]{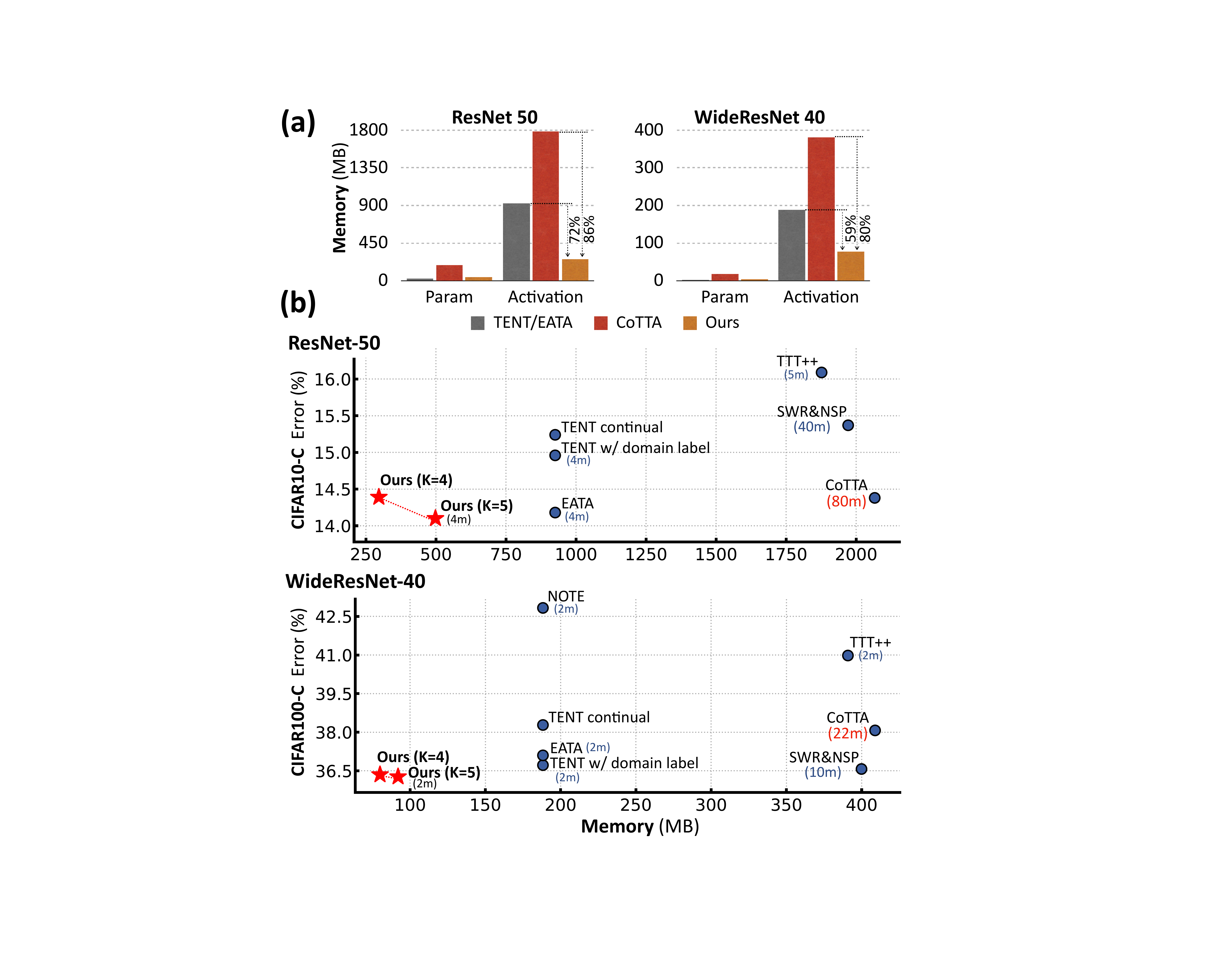}
\vspace{-1.5em}
\caption{\textbf{(a) Memory cost comparison between TTA methods}. The size of activations, not the parameters, is the primary memory bottleneck during training. 
\textbf{(b) CIFAR-C adaptation performance.} We perform the continual online adaptation on CIFAR-C dataset. The x- and y-axis are the average error of all corruptions and the total memory consumption including the parameters and activations, respectively. Our approach, EcoTTA, achieves the best results while consuming the least amount of memory, where K is the model partition factor used in our method.}
\label{fig:figure1}
\vspace{-1.5em}
\end{figure}

Despite recent advances in deep learning~\cite{dosovitskiy2020image,resnet,he2020momentum,he2022masked}, deep neural networks often suffer from performance degradation when the source and target domains differ significantly~\cite{robusetnet,long2016unsupervised, li2021semantic}. 
Among several tasks addressing such domain shifts, test-time adaptation (TTA) has recently received a significant amount of attention due to its practicality and wide applicability especially in on-device settings~\cite{tent, ttt++, iwasawa2021test, gandelsman2022test}.
This task focuses on adapting the model to unlabeled online data from the target domain without access to the source data.

\begin{figure*}[t]\centering
\vspace{-0.5em}
\includegraphics[width=1.0\linewidth]{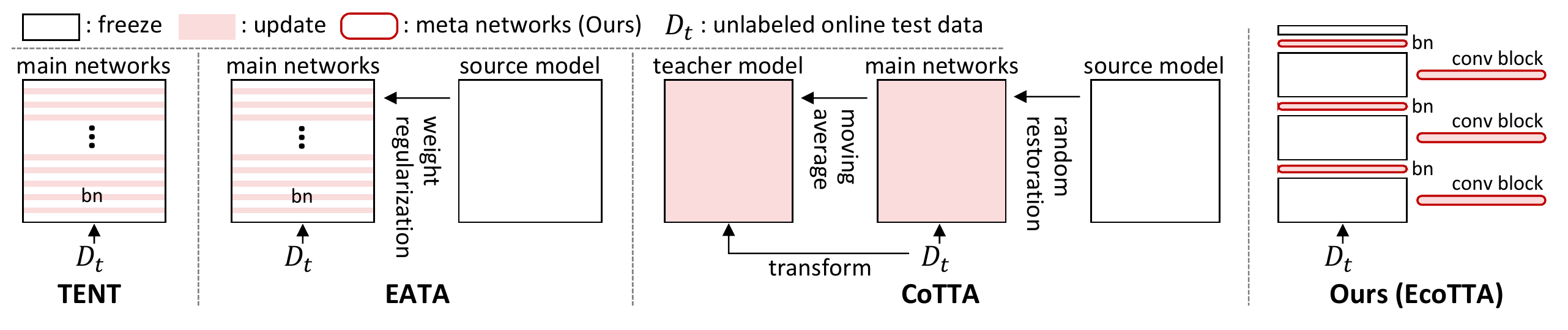}\vspace{-.7em}
\caption{\textbf{Architecture for test-time adaptation.} We illustrate TTA methods: TENT~\cite{tent}, EATA~\cite{eata}, CoTTA~\cite{cotta}, and Ours (EcoTTA). TENT and EATA update \textit{multiple} batch norm layers, in which large activations have to be stored for gradient calculation. In CoTTA, an entire network is trained with additional strategies for continual adaptation that requires a significant amount of both memory and time. In contrast, our approach requires a minimum size of activations by updating only \textit{a few} layers. Also, stable long-term adaptation is performed by our proposed regularization, named self-distilled regularization.} 
\label{fig:frameworks}
\vspace{-1.5em}
\end{figure*}

While existing TTA methods show improved TTA performances, minimizing the sizes of memory resources have been relatively under-explored, which is crucial considering the applicability of TTA in on-device settings.
For example, several studies~\cite{cotta,ttt++,swr} update entire model parameters to achieve large performance improvements, which may be impractical when the available memory sizes are limited. 
Meanwhile, several TTA approaches update only the batch normalization (BN) parameters~\cite{tent, eata, note} to make the optimization efficient and stable
However, even updating only BN parameters is not memory efficient enough since the amount of memory required for training models significantly depends on the size of intermediate activations rather than the learnable parameters~\cite{tinytl,dhar2021survey,repnet}. 
Throughout the paper, activations refer to the intermediate features stored during the forward propagation, which are used for gradient calculations during backpropagation.
\figref{fig:figure1} (a) demonstrates such an issue. 

Moreover, a non-trivial number of TTA studies assume a stationary target domain~\cite{tent, ttt++, swr, shin2022mm}, but the target domain may continuously change in the real world (\textit{e.g.,} continuous changes in weather conditions, illuminations, and location~\cite{robusetnet} in autonomous driving). Therefore, it is necessary to consider long-term TTA in an environment where the target domain constantly varies.
However, there exist two challenging issues: 1) catastrophic forgetting~\cite{cotta, eata} and 2) error accumulation.
Catastrophic forgetting refers to degraded performance on the source domain due to long-term adaptation to target domains~\cite{cotta, eata}.
Such an issue is important since the test samples in the real world may come from diverse domains, including the source and target domains~\cite{eata}.
Also, since target labels are unavailable, TTA relies on noisy unsupervised losses, such as entropy minimization~\cite{entmin}, so long-term continual TTA may lead to error accumulation~\cite{erroraccum, confirmbias}.

To address these challenges, we propose memory-\textbf{E}fficient \textbf{co}ntinual \textbf{T}est-\textbf{T}ime \textbf{A}daptation (EcoTTA), a simple yet effective approach for 1) enhancing memory efficiency and 2) preventing catastrophic forgetting and error accumulation. First, we present a memory-efficient architecture consisting of frozen original networks and our proposed meta networks attached to the original ones. 
During the test time, we freeze the original networks to discard the intermediate activations that occupy a significant amount of memory. 
Instead, we only adapt lightweight meta networks to the target domain, composed of only one batch normalization and one convolution block.
Surprisingly, updating only the meta networks, not the original ones, can result in significant performance improvement as well as considerable memory savings. Moreover, we propose a self-distilled regularization method to prevent catastrophic forgetting and error accumulation. 
Our regularization leverages the preserved source knowledge distilled from the frozen original networks to regularize the meta networks. Specifically, we control the output of the meta networks not to deviate from the one extracted by the original networks significantly. 
Notably, our regularization leads to negligible overhead because it requires no extra memory and is performed in parallel with adaptation loss, such as entropy minimization.

Recent TTA studies require access to the source data \emph{before model deployments}~\cite{ttt++,swr,cafa,adachi22cafa,lim2023ttn,eata}.
Similarly, our method uses the source data to warm up the newly attached meta networks for a small number of epochs before model deployment. 
If the source dataset is publicly available or the owner of the pre-trained model tries to adapt the model to a target domain, access to the source data is feasible~\cite{swr}. 
Here, we emphasize that pre-trained original networks are frozen throughout our process, and our method is applicable to any pre-trained model because it is agnostic to the architecture and pre-training method of the original networks.

Our paper presents the following contributions:
\vspace{-.5em}
\begin{itemize}
    \setlength\itemsep{-.3em}
    \item We present novel meta networks that help the frozen original networks adapt to the target domain. This architecture significantly minimize memory consumption up to 86\% by reducing the activation sizes of the original networks.
    \item We propose a self-distilled regularization that controls the output of meta networks by leveraging the output of frozen original networks to preserve the source knowledge and prevent error accumulation.
    \item We improve both memory efficiency and TTA performance compared to existing state-of-the-art methods on 1) image classification task (\eg., CIFAR10/100-C and ImageNet-C) and 2) semantic segmentation task (\eg., Cityscapes with weather corruption)
\end{itemize}

%% file: section/related.tex
\begin{figure*}[t]\centering
\includegraphics[width=1.\linewidth]{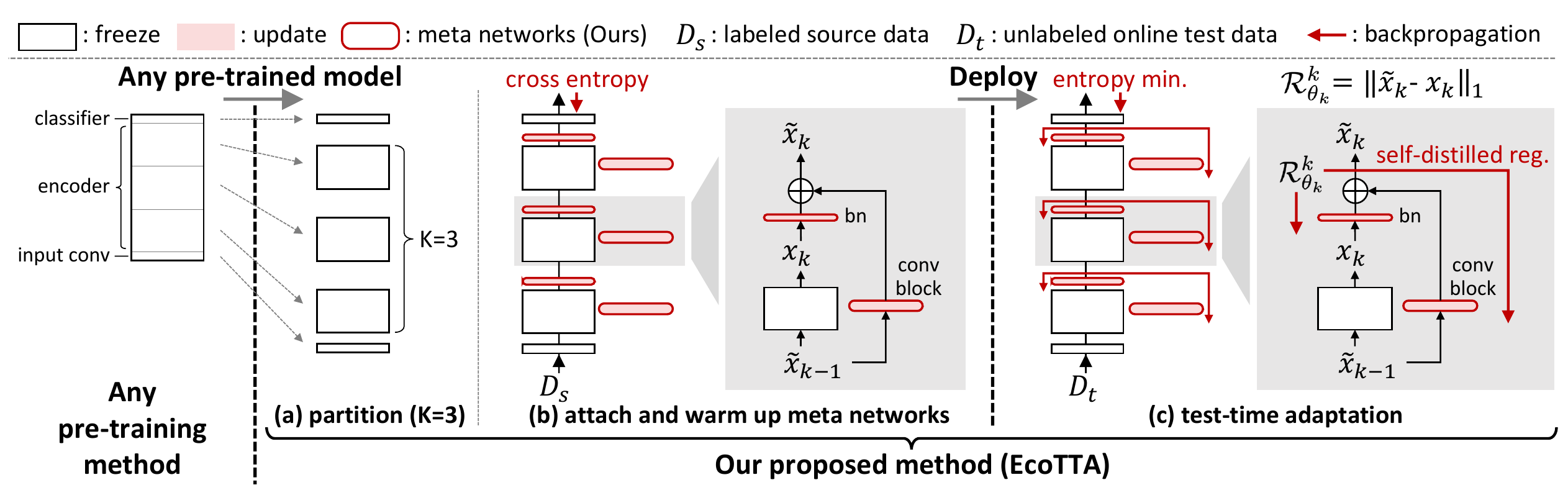}\vspace{-.5em}
\caption{\textbf{Overview of our approach.} 
\textbf{(a)} The encoder of the pre-trained model is divided into K parts (\textit{i.e.,} model partition factor K). \textbf{(b)} Before deployment, the meta networks are attached to each part of the original networks and pre-trained with source dataset $\mathcal{D}_s$. \textbf{(c)} After the model is deployed, \textit{only} the meta networks are updated with unsupervised loss (\ie, entropy minimization) on target data $\mathcal{D}_t$, while the original networks are frozen. To avoid error accumulation and catastrophic forgetting by the long-term adaptation, we regularize the output $\tilde{x}_{k}$ of each group of the meta networks leveraging the output $x_{k}$ of the \textit{frozen} original network, which preserves the source knowledge.}
\label{fig:approach}\vspace{-1.5em}
\end{figure*}

\paragraph{Mitigating domain shift.} One of the fundamental issues of DNNs is the performance degradation due to the domain shift between the train~(\ie source) and test~(\ie target) distributions. Several research fields attempt to address this problem, such as unsupervised domain adaptation~\cite{advent, adversarial_da, park2020discover, shin2021labor, IAST, twophase} and domain generalization~\cite{zhou2021mixstyle,robusetnet}. In particular, domain generalization aims to learn invariant representation so as to cover the possible shifts of test data. They simulate the possible shifts using a single or multiple source dataset~\cite{zhou2021mixstyle, zhang2022exact, li2022uncertainty} or force to minimize the dependence on style information~\cite{pan2018two, robusetnet}. However, it is challenging to handle all potential test shifts using the given source datasets~\cite{gulrajani2020search}. 
Thus, instead of enhancing generalization ability during the training time, TTA~\cite{tent} overcomes the domain shift by directly adapting to the test data.

\paragraph{Test-time adaptation.} Test-time adaptation allows the model to adapt to the test data (\ie, target domain) in a source-free and online manner~\cite{5995317,sun2020testtime,tent}.
Existing works improve TTA performance with sophisticated designs of unsupervised loss~\cite{confi_max,zhang2021memo,ttt++, swr, marsden2022gradual, shin2022mm, chen2022contrastive, gandelsman2022test, adachi22cafa, azarian2023test, das2023transadapt, song2022cd} or enhance the usability of small batch sizes~\cite{khurana2021sita, abn, hu2021mixnorm, wildTTA, lim2023ttn} considering streaming test data. They focus on improving the adaptation performance with a stationary target domain (\ie, single domain TTA setup). 
In such a setting, the model that finished adaptation to a given target domain is reset to the original model pre-trained with the source domain in order to adapt to the next target domain.

Recently, CoTTA~\cite{cotta} has proposed continual TTA setup to address TTA under a continuously changing target domain which also involves a long-term adaptation. This setup frequently suffers from error accumulation~\cite{erroraccum, confirmbias, confirmbias2} and catastrophic forgetting~\cite{cotta, eval_cotta,eata}.
Specifically, performing a long-term adaptation exposes the model to unsupervised loss from unlabeled test data for a long time, so errors are accumulated significantly.
Also, the model focuses on learning new knowledge and forgets about the source knowledge, which becomes problematic when the model needs to correctly classify the test sample as similar to the source distribution. 
To address such issues, CoTTA~\cite{cotta} randomly restores the updated parameters to the source one, while EATA~\cite{eata} proposed a weight regularization loss.

\paragraph{Efficient on-device learning.} Since the edge device is likely to be memory constrained (\eg, a Raspberry Pi with 512MB and iPhone 13 with 4GB), it is necessary to take account of the memory usage when deploying the models on the device~\cite{lin2020mcunet}. 
TinyTL~\cite{tinytl}, a seminal work in on-device learning, shows that the activation size, not learnable parameters, bottlenecks the training memory.
Following this, recent on-device learning studies~\cite{tinytl, yang2022da3,repnet} targeting fine-tuning task attempt to decrease the size of intermediate activations.
In contrast, previous TTA studies~\cite{tent,eata} have overlooked these facts and instead focused on reducing learnable parameters.
This paper, therefore, proposes a method that not only reduces the high activation sizes required for TTA, but also improves adaptation performance.

%% file: section/approach.tex
\figref{fig:approach} illustrates our simple yet effective approach which only updates the newly added meta networks on the target domain while regularizing them with the knowledge distilled from the frozen original network. 
This section describes how such a design promotes memory efficiency and prevents error accumulation and catastrophic forgetting which are frequently observed in long-term adaptation.

\subsection{Memory-efficient Architecture}
\label{sec:approach1}

\paragraph{Prerequisite.} 
We first formulate the forward and the backward propagation.
Assume that the $i^{th}$ linear layer in the model consists of weight $\mathcal{W}$ and bias~$b$, and the input and output features of this layer are $f_{i}$ and $f_{i+1}$, respectively. 
Given that the forward propagation of $f_{i+1} = f_{i}\mathcal{W} + b$, the backward propagation from the $i{+}1$$^{th}$ layer to the $i^{th}$ layer, and the weight gradient are respectively formulated as:
\begin{equation}
\label{eq:backward}
{\footnotesize
    \frac{\partial \loss}{\partial f_{i}}= \frac{\partial \loss}{\partial f_{i+1}} \mathcal{W}^{T}, \quad
    \frac{\partial \loss}{\partial \mathcal{W}}= \textcolor{darkred}{f_i^{T}} \frac{\partial \loss}{\partial f_{i+1}}.
    }
\end{equation}

\noindent
\equref{eq:backward} means that the learnable layers whose weight $\mathcal{W}$ need to be updated must store intermediate activations $\textcolor{darkred}{f_i}$ to compute the weight gradient. In contrast, the backward propagation in frozen layers can be accomplished without saving the activations, only requiring its weight $\mathcal{W}$. Further descriptions are provided in Appendix A.

TinyTL~\cite{tinytl} shows that activations occupy the majority of the memory required for training the model rather than learnable parameters. 
Due to this fact, updating the entire model (e.g., CoTTA~\cite{cotta}) requires a substantial amount of memory.
Also, updating only parameters in batch normalization (BN) layers (e.g., TENT~\cite{tent} and EATA~\cite{eata}) is not an effective approach enough since they still save the large intermediate activations for \textit{multiple} BN layers. 
While previous studies fail to reduce memory by utilizing large activations, this work proposes a simple yet effective way to reduce a significant amount of memory by discarding them. 

\paragraph{Before deployment.} 
\label{sec:before_deployment}
As illustrated in \figref{fig:approach} (a, b), we first take a pre-trained model using any pre-training method. 
We divide the encoder of the pre-trained model into K number of parts and attach lightweight meta networks to each part of the original network. 
The details of how to divide the model into K number of parts are explained in the next section.
One group of meta network composes of one batch normalization layer and one convolution block (\ie, Conv-BN-Relu). 
Before the deployment, we pre-train the meta networks on the source dataset $\mathcal{D}_s$ for a small number of epochs (e.g., 10 epochs for CIFAR dataset) while freezing the original networks.
Such a warm-up process is completed before the model deployment, similarly done in several TTA works~\cite{swr,cafa,lim2023ttn,eata}.
Note that we do not require source dataset $\mathcal{D}_s$ during test time.

\paragraph{Pre-trained model partition.} 
Previous TTA studies addressing domain shifts~\cite{swr, confi_max} indicate that updating shallow layers is more crucial for improving the adaptation performance than updating the deep layers. 
Inspired by such a finding, given that the encoder of the pre-trained model is split into model partition factor K (\eg, 4 or 5), we partition the shallow parts of the encoder more (\ie., densely) compared to the deep parts of it. Table~\ref{tab:model_partition} shows how performance changes as we vary the model partition factor K.

\paragraph{After deployment.} 
During the test-time adaptation, we only adapt meta networks to target domains while freezing the original networks. Following EATA~\cite{eata}, we use the entropy minimization $H(\hat{y})=-\sum_c p(\hat{y}) \log p(\hat{y})$ to the samples achieving entropy less than the pre-defined entropy threshold $H_0$, where $\hat{y}$ is the prediction output of a test image from test dataset $\mathcal{D}_t$ and p($\cdot$) is the softmax function. Thus, the main task loss for adaptation is defined as 
\begin{equation}
\label{eq:entloss}
    \loss^{ent} = \mathbb{I}_{\{H(\hat{y}) < H_0\}} 	\cdot H(\hat{y}),
\end{equation}
\noindent
where $\mathbb{I}_{\{\cdot\}}$ is an indicator function. In addition, in order to prevent catastrophic forgetting and error accumulation, we apply our proposed regularization loss $\mathcal{R}^{k}$, which is described next in detail. Consequently, the overall loss of our method is formulated as,
\begin{equation}
\label{eq:totalloss}
    \loss^{total}_{\theta} = \loss_{\theta}^{ent} + \lambda \; \sum_{k}^{K} \mathcal{R}_{\theta_k}^{k},
\end{equation}
\noindent
where $\theta $ and $\theta_k$ denotes parameters of all meta networks and those of k-th group of meta networks, respectively, and $\lambda$ is used to balance the scale of the two loss functions. Note that our architecture requires less memory than previous works~\cite{cotta, tent} since we use frozen original networks and discard its intermediate activations. To be more specific, our architecture uses 82\% and 60\% less memory on average than CoTTA and TENT/EATA.

\subsection{Self-distilled Regularization}
\label{sec:approach2}

\input{table/main_cifar}
\input{table/main_imagenet}
\input{table/main_odl}

The unsupervised loss from unlabeled test data $\mathcal{D}_t$ is likely to provide a false signal (\ie, noise) to the model ($\hat{y} \neq y_t$ where $y_t$ is the ground truth test label). 
Previous works have verified that long-term adaptation with unsupervised loss causes overfitting due to error accumulation~\cite{erroraccum, confirmbias} and catastrophic forgetting~\cite{cotta, eval_cotta}. 
To prevent the critical issues, we propose a self-distilled regularization utilizing our architecture. 
As shown in Fig.~\ref{fig:approach}, we regularize the output $\tilde{x}_k$ of each $k$-th group of the meta networks not to deviate from the output $x_k$ of the $k$-th part of frozen original networks. 
Our regularization loss which computes the mean absolute error (\ie, L1 loss) is formulated as follows:
\begin{equation}
\label{eq:regloss}
    \mathcal{R}_{\theta_k}^{k} = \left\|\tilde{x}_k-x_k\right\|_1 .
\end{equation}
\noindent
Since the original networks are not updated, the output $x_{k, k \sim  K}$ extracted from them can be considered as containing the knowledge learned from the \textit{source} domain. 
Taking advantage of this fact, we let the output of meta networks $\tilde{x}_{k}$ be regularized with knowledge distilled from the original networks.
By preventing the adapted model to not significantly deviate from the original model, we can prevent 1) catastrophic forgetting by maintaining the source domain knowledge and 2) error accumulation by utilizing the class discriminability of the original model.
Remarkably, unlike previous works~\cite{cotta, eata}, our regularization does not require saving additional original networks, which accompanies extra memory usage.
Moreover, it only needs a negligible amount of computational overhead because it is performed in parallel with the entropy minimization loss $\loss^{ent}$.

%% file: table/main_cifar.tex
\begin{table*}[t]
\centering
\subfloat[
\textbf{CIFAR10-C with severity level 5}
\label{tab:cifar10}
]{
\centering
\begin{minipage}{0.565\linewidth}{\begin{center}

{\renewcommand{\arraystretch}{1.2}
\resizebox{\textwidth}{!}{
{\Large
\begin{tabular}{lcccccc} 
\toprule
                                               & \multicolumn{2}{c}{\textbf{WideResNet-40 (AugMix)}}  & \multicolumn{2}{c}{\textbf{WideResNet-28}}      & \multicolumn{2}{c}{\textbf{ResNet-50}}  \\
\textbf{Method}                                         & \textbf{Avg. err {\scriptsize $\downarrow$}}        & \textbf{Mem. (MB)}                                                                           & \textbf{Avg. err {\scriptsize $\downarrow$}}  & \textbf{Mem. (MB)}                                                                           & \textbf{Avg. err {\scriptsize $\downarrow$}} & \textbf{Mem. (MB)}                                                                            \\ 
\drule
Source                                         & 36.7            & 11                                                                                              & 43.5      & 58                                                                                             & 48.8     & 91                                                                                              \\
BN Stats Adapt~\cite{tbn}                                            & 15.4            & 11                                                                                              & 20.9      & 58                                                                                             & 16.6     & 91                                                                                              \\ 
\hline
Single do. TENT~\cite{tent}                                    & 12.7            & 188                                                                                            & 19.2      & 646                                                                                            & 15.0     & 925                                                                                             \\
Continual TENT                                 & 13.3            & 188                                                                                            & 20.0      & 646                                                                                            & 15.2     & 925                                                                                             \\
TTT++~\cite{ttt++}                                          & 14.6            & 391                                                                                            & 20.3      & 1405                                                                                           & 16.1     & 1877                                                                                            \\
SWR\&NSP~\cite{swr}                                         & \underbar{12.1}            & 400                                                                                            & 17.2      & 1551                                                                                           & 15.4     & 1971                                                                                            \\
NOTE~\cite{note}                                           & 13.4            & 188                                                                                            & 20.2      & 646                                                                                            & -        & -                                                                                               \\
EATA~\cite{eata}                                           & 13.0            & 188                                                                                            & 18.6      & 646                                                                                            & \underbar{14.2}     & 925                                                                                             \\
CoTTA~\cite{cotta}                                          & 14.0            & 409                                                                                            & 17.0      & 1697                                                                                           & 14.4     & 2066                                                                                            \\
\rowcolor[rgb]{0.9,0.9,0.9} \textbf{Ours (K=4)} & 12.2            & 80~{\small (80, 58\%}{\scriptsize $\downarrow$}{\small )}                                                                                             & \underbar{16.9}      & 404~{\small (76, 38\%}{\scriptsize $\downarrow$}{\small )}                                                                                            & 14.4     & 296~{\small (86, 68\%}{\scriptsize $\downarrow$}{\small )}                                                                                             \\

\rowcolor[rgb]{0.9,0.9,0.9} \textbf{Ours (K=5)}  & \textbf{12.1}            & 92~{\small (77, 51\%}{\scriptsize $\downarrow$}{\small )}                                                                                             & \textbf{16.8}      & 471~{\small (72, 27\%}{\scriptsize $\downarrow$}{\small )}                                                                                            & \textbf{14.1}     & 498~{\small (76, 46\%}{\scriptsize $\downarrow$}{\small )}                                                                                             \\
\toprule
\end{tabular}}}}

\end{center}

}\end{minipage}
}
\subfloat[
\textbf{CIFAR100-C with severity level 5}
\label{tab:cifar100}
]{
\begin{minipage}{0.425\linewidth}{\begin{center}

{\renewcommand{\arraystretch}{1.2}
\resizebox{\textwidth}{!}{
{\Large
\begin{tabular}{lcccc} 
\toprule
                                             & \multicolumn{2}{c}{\textbf{WideResNet-40 (AugMix)}} & \multicolumn{2}{c}{\textbf{ResNet-50}}  \\
\textbf{Method}                                       & \textbf{Avg. err {\scriptsize $\downarrow$}} & \textbf{Mem. (MB)}                      & \textbf{Avg. err {\scriptsize $\downarrow$}} & \textbf{Mem. (MB)}          \\ 
\drule
Source                                       & 69.7     & 11                              & 73.8     & 91                 \\
BN Stats Adapt~\cite{tbn}                                          & 41.1     & 11                              & 44.5     & 91                 \\ 
\hline
Single do. TENT~\cite{tent}                                  & 36.7     & 188                            & 40.1     & 926                \\
Continual TENT                               & 38.3     & 188                            & 45.9     & 926                \\
TTT++~\cite{ttt++}                                        & 41.0     & 391                            & 44.2     & 1876               \\
SWR\&NSP~\cite{swr}                                       & 36.6     & 400                            & 44.1     & 1970               \\
NOTE~\cite{note}                                         & 42.8     & 188                            & -        & -                  \\
EATA~\cite{eata}                                         & 37.1     & 188                            & 39.9     & 926                \\
CoTTA~\cite{cotta}                                        & 38.1     & 409                            & 40.2     & 2064               \\
\rowcolor[rgb]{0.9,0.9,0.9} \textbf{Ours (K=4)} & \underbar{36.4}     & 80~{\small (80, 58\%}{\scriptsize $\downarrow$}{\small )}                             & \underbar{39.5}     & 296~{\small (86, 68\%}{\scriptsize $\downarrow$}{\small )}                \\
\rowcolor[rgb]{0.9,0.9,0.9} \textbf{Ours (K=5)} & \textbf{36.3}     & 92~{\small (77, 51\%}{\scriptsize $\downarrow$}{\small )}                             & \textbf{39.3}     & 498~{\small (76, 46\%}{\scriptsize $\downarrow$}{\small )}                \\
\toprule
\end{tabular}}}}
\end{center}}

\end{minipage}
}

\vspace{-.7em}
\caption{\textbf{Comparison of error rate ($\%$) on CIFAR-C.} We report an average error of 15 corruptions on \textit{continual} TTA and a memory requirement including model parameters and activation sizes. The lowest error is in bold, and the second lowest error is underlined. The memory reduction rates compared to CoTTA and TENT are presented sequentially. 
WideResNet-40 was pre-trained with AugMix~\cite{augmix} that is a data processing to increase the robustness of the model.
Source denotes the pre-trained model without adaptation. Single domain (in short, single do.) TENT resets the model when adapting to a new target domain, so the domain labeles are required.}
\label{tab:main_cifar} 
\vspace{-1.4em}
\end{table*}

%% file: table/main_imagenet.tex
{\renewcommand{\arraystretch}{1.2}
\begin{table}
\centering
\resizebox{.46\textwidth}{!}{
{\Large
\begin{tabular}{l c c c } 
\toprule
                                              & \textbf{ResNet-50 (AugMix)}                 & \textbf{ResNet-50}            & \multicolumn{1}{c}{\multirow{2}{*}{\begin{tabular}[c]{@{}c@{}}(MB)\\\textbf{Total Mem. {\scriptsize $\downarrow$}}\end{tabular}}}                \\
\textbf{Method}  & $\qquad $\textbf{Avg. err {\scriptsize $\downarrow$}}$\qquad $                      & $\qquad $\textbf{Avg. err {\scriptsize $\downarrow$}}$\qquad $                  & \multicolumn{1}{c}{}   \\ 
\drule
Source                                        & 74.36                                                       & 82.35      & 91                                                                                               \\
BN Stats Adapt~\cite{tbn}                                      & 57.87                                                       & 72.18      & 91                                                                                             \\
Continual TENT~\cite{tent}                                          & 56.1 (0.6)                                                  & 66.2 (1.1) & 1486                                                                                    \\
EATA~\cite{eata}                                          & 54.9 (2.3)                                                  & 63.8 (2.7) & 1486                                                                             \\
CoTTA~\cite{cotta}                                         & \underbar{54.6} (3.9)                                       & \textbf{62.6} (3.1) & 3132                                                                            \\
\rowcolor[rgb]{0.9,0.9,0.9} \textbf{Ours (K=4)} & 55.2 (3.0)                                                  & 64.6 (3.2) & \textbf{438} {\small (86, 72\%}{\scriptsize $\downarrow$}{\small )}                                                                                        \\
\rowcolor[rgb]{0.9,0.9,0.9} \textbf{Ours (K=5)} & \textbf{54.4} (2.7)                                         & \underbar{63.4} (3.0) & \underbar{747} {\small (75, 51\%}{\scriptsize $\downarrow$}{\small )}                                                                                     \\
\toprule
\end{tabular}}}
\small
\vspace{-0.2cm}
\caption{\textbf{Comparison of error rate ($\%$) on ImageNet-C with severity level 5.} Standard deviation for ten diverse corruption sequences is denoted by the parentheses values. 
The total memory refers to the sum of model parameters and activations.}
\label{tab:main_imagenet} 
\vspace{-0.5em}
\end{table}}

%% file: table/main_odl.tex
{\renewcommand{\arraystretch}{1.2}
\begin{table}
\centering
\resizebox{.47\textwidth}{!}{
{\Large
\begin{tabular}{lccccc} 
\toprule
Avg. err (\%)       &   & \multicolumn{2}{c}{\textbf{CIFAR10-C}}      & \multicolumn{2}{c}{\textbf{CIFAR100-C}}      \\
\textbf{Method}                                 & \textbf{Mem. (MB)} & single do. & continual & single do. & continual  \\ 
\drule
BN Stats Adapt~\cite{tbn}                                 & 91           & 16.6                 &16.6 & 44.5      &  44.5\\
TinyTL$^{\dag}$~\cite{tinytl}                                 & 379           & 15.8                 & 21.9      & \underbar{40.5}                 & 77.4       \\
RepNet$^{\dag}$~\cite{repnet}                                 & 508           & \underbar{15.2}                 & 20.9      & 41.5                 & 52.1           \\
AuxAdapt$^{\dag}$~\cite{auxadapt}                             & \textbf{207}  & 16.0                 & \underbar{16.7}          & 44.0                 & \underbar{45.8}            \\
\rowcolor[rgb]{0.9,0.9,0.9} \textbf{Ours (K=4)} & \underbar{296}         & \textbf{14.4}                 & \textbf{14.4}      & \textbf{39.5}                 & \textbf{39.2}       \\
\bottomrule
\end{tabular}}}
\footnotesize
\vspace{-0.2cm}
\caption{\textbf{Comparison with methods for on-device learning.} The backbone is ResNet-50. $^{\dag}$ denotes our own re-implemented models. single do. indicates the singe domain TTA setup.}
\label{tab:main_odl} 
\vspace{-1.8em}
\end{table}}

%% file: section/experiment1.tex
\input{table/ablation_arch}

We evaluate our approach to image classification tasks based on the continual test-time adaptation setup with three datasets: CIFAR10-C, CIFAR100-C, and ImageNet-C.

\paragraph{Experimental setup.} Following CoTTA~\cite{cotta}, we conduct most experiments on the continual TTA task, where we continually adapt the deployed model to each corruption type sequentially without resetting the model. This task is more challenging but more realistic than single domain TTA task~\cite{tent} in which the adapted model is periodically reset to the original pre-trained model after finishing adaptation to each target, so they require additional domain information. 
Moreover, we evaluate our approach on the long-term TTA setup, which is detailed in \secref{sec:empirical}.

Following the previous TTA studies~\cite{tent,cotta}, we evaluate models with \{CIFAR10, CIFAR10-C\}, \{CIFAR100, CIFAR100-C\}, and \{ImageNet, ImageNet-C\} where the first and the second dataset in each bracket refers to the source and the target domain, respectively. 
The target domains include 15 types of corruptions (\eg noise, blur, weather, and digital) with 5 levels of severity, which are widely used in conventional benchmarks~\cite{imagenetC}.

\paragraph{Implementation Details.} We evaluate our approach within the frameworks officially provided by previous state-of-the-art methods~\cite{cotta, eata}. 
For fair comparisons, we use the same pre-trained model, which are WideResNet-28 and WideResNet-40~\cite{wrn} models from the RobustBench~\cite{robustbench}, and ResNet-50~\cite{resnet} model from TTT++~\cite{ttt++, swr}. 
Before the deployment, we pre-train the meta networks on the source dataset using a cross-entropy loss with SGD optimizer with the learning rate of 5e-2. 
Since the meta networks contain only a few layers, we pre-train them with a small number of epochs: 10 and 3 epochs for CIFAR and ImageNet, respectively.
After deployment, similar to EATA~\cite{eata},
we use the same SGD optimizer with the learning rate of 5e-3. 
In \equref{eq:entloss}, the entropy threshold $H_0$ is set to $0.4 \times \ln C$ where $C$ denotes the number of task classes. 
The batch size is 64 and 32 for CIFAR and ImageNet, respectively. 
We set the importance of the regularization $\lambda$ in \equref{eq:totalloss} to 0.5 to balance it with the entropy minimization loss. Additional implementation details can be found in Appendix C.

\paragraph{Evaluation Metric.} For all the experiments, we report error rates calculated during testing and the memory consumption, including the model parameter and the activation storage. We demonstrate the memory efficiency of our work by using the official code provided by TinyTL~\cite{tinytl}.

\subsection{Comparisons}\label{sec:comparisons}

\paragraph{Comparisons with TTA methods.} 
We compare our approach to competing TTA methods on extensive benchmarks and various pre-trained models. The results of CIFAR10/100-C are detailed in \tabref{tab:main_cifar}. The model partition factor K are set to 4 and 5. Our approach outperforms existing TTA methods with the lowest memory usage in all pre-trained models. Specifically, in WideResNet-40, our method achieves superior performance while requiring 80\% and 58\% less memory than CoTTA~\cite{cotta} and EATA~\cite{eata}, respectively, which are also designed for continual TTA. Approaches targeting single domain TTA~\cite{tent, ttt++, swr} show poor performance due to error accumulation and catastrophic forgetting, as observed in CoTTA. The error rates for each corruption type are provided in Appendix F.

\tabref{tab:main_imagenet} shows the experiment for ImageNet-C. Two ResNet-50 backbones from RobustBench~\cite{robustbench} are leveraged. Following CoTTA, evaluations are conducted on ten diverse corruption-type sequences. We achieve comparable performance to CoTTA while utilizing 86\% and 75\% less memory with K=4 and 5, respectively. 
In addition, we observe that our approach shows superior performance when adopting the model pre-trained with strong data augmentation methods (\textit{e.g.,} Augmix~\cite{augmix}).

\input{table/ablation_k_memory}

\input{table/ablation_regular}

\paragraph{Comparisons with on-device learning methods.} 
We compare our approach with methods for memory-efficient on-device learning. TinyTL~\cite{tinytl} and RepNet~\cite{repnet} focus on \emph{supervised} on-device learning~(\ie, requiring labeled target data). 
However, since TTA assumes that we do not have access to the target labels, utilizing such methods to TTA directly is infeasible.
Therefore, we experimented by replacing supervised loss~(\ie, cross-entropy) with unsupervised loss~(\ie, entropy minimization) in TinyTL and RepNet. 
As shown in \tabref{tab:main_odl}, they suffer from performance degradation in continual TTA, showing inferior performance compared to our proposed approach even in the single domain TTA.

Similar to ours, AuxAdapt~\cite{auxadapt} adds and updates a small network (\ie, ResNet-18) while freezing the pre-trained model. 
Unlike our approach, they only modify a prediction output, not intermediate features. While AuxAdapt requires the least memory usage, it fails to improve TTA performan-ce in single domain TTA. Nevertheless, since the original model is frozen, it suffers less from catastrophic forgetting and error accumulation than TinyTL~\cite{tinytl} and RepNet~\cite{repnet} in the continual TTA.
Through the results, we confirm that our proposed method brings both memory efficiency and a significant performance improvement in both TTA setups.

\subsection{Empirical Study}
\label{sec:empirical}

\paragraph{Architecture design.} 
An important design of our meta networks is injecting a single BN layer before the original networks and utilizing a residual connection with one conv block.
\tabref{tab:auxnet_ablation} studies the effectiveness of the proposed design by comparing it with six different variants.
From the results, we observe that using only either conv block (ii) or BN (iii) aggravates the performance: error rate increases by 1.4\% and 3.8\% on CIFAR-100-C with WideResNet-40. 

In design (i), we enforce both BN parameters and Conv layers in the meta networks to take the output of the original networks as inputs.
Such a design brings performance drop. We speculate that it is because the original network, which is not adapted to the target domain, lacks the ability to extract sufficiently meaningful features from the target image.
Also, we observed a significant performance degradation after removing the residual connection in design (iv).
In addition, since attention mechanisms~\cite{cbam, senet} generally have improved classification accuracy, we study how attention mechanisms can further boost TTA performance of our approach in design (v, vi).
The results show that it is difficult for the attention module to train ideally in TTA setup using unsupervised learning, unlike when applying it to supervised learning. An ablation study on each element of meta networks can be found in Appendix D.

\input{table/ablation_smallbatch}

\paragraph{Number of blocks in each partition.} 
ResNet~\cite{resnet} consists of multiple residual blocks (\eg, BasicBlock and Bottleneck in Pytorch~\cite{pytorch}).  For instance, WideResNet-28 has 12 residual blocks. 
By varying the number of blocks for each part of the original networks, we analyze TTA performance in \tabref{tab:model_partition}.  
We observe that splitting the shallow parts of the encoder densely (\eg, 2,2,4,4 blocks, from the shallow to the deep parts sequentially) brings more performance gain than splitting the deep layers densely (\eg, 4,4,2,2 blocks). 
We suggest that it is because we modify the lower-level feature more as we split shallow layers densely. 
Our observation is aligned with the finding of previous TTA works~\cite{swr, confi_max}, which show that updating the shallow layers more than the deep layers improves TTA performance.

\paragraph{Number of model partition K.} 
\figref{fig:k_memory} shows both memory requirement and adaptation performance according to the model partition factor K. With a small K (\eg, 1 or 2), the intermediate outputs are barely modified, making it difficult to achieve a reasonable level of performance.
We achieve the best TTA performance with K of 4 or 5 as adjusting a greater numver of intermediate features.
In the meanwhile, we observe that the average error rate is saturated and remains consistent when K is set to large values (\eg 6,7 or 8) even with the increased amount of activations and learnable parameters. Therefore, we set K to 4 and 5.

\input{table/main_segment}

\paragraph{Catastrophic forgetting.}
We conduct experiments to confirm the catastrophic forgetting effect (\figref{fig:forgetting}). 
Once finishing adaptation to each corruption, we evaluate the model on clean target data (\ie, test-set of CIFAR dataset) without updating the model. 
For TENT with no regularization, the error rates for the clean target data (\ie, clean error (\%)) increase gradually, which can be seen as the phenomenon of catastrophic forgetting. 
In contrast, our approach consistently maintains the error rates for the clean target data, proving that our regularization loss effectively prevents catastrophic forgetting. 
These results indicate that our method can be reliably utilized in various domains, including the source and target domains. 

\paragraph{Error accumulation in long-term adaptation.} 
To evaluate the error accumulation effect, we repeat all the corruption sequences for 100 rounds. The results are described in \figref{fig:long}.
For TENT, a gradual increase in error rates is observed in later rounds, even with small learning rates. 
For example, TENT~\cite{tent} with the learning rate of 1e-5 achieves the error rate of 39.7\%, and reached its lowest error rate of 36.5\% after 8 rounds. 
However, it shows increased error rate of 38.6\% after 100 rounds due to overfitting.
It suggests that without regularization, TTA methods eventually face overfitting in long-term adaptation~\cite{erroraccum,confirmbias,eval_cotta}.
Our method in the absence of regularization ($\lambda = 0$) also causes overfitting. 
On the other hand, when self-distilled regularization is involved ($\lambda > 0$), the performance remains consistent even in the long-term adaptation. 

\paragraph{Small batch size.} 
We examine the scalability of our approach with a TTA method designed for small batches size, named adapting BN statistics (\ie, AdaptBN~\cite{adapt_abn,zhang2021memo}). When the number of batches is too small, the estimated statistics can be unreliable~\cite{adapt_abn}. Thus, they calibrate the source and target statistics for the normalization of BN layers so as to alleviate the domain shift and preserve the discriminative structures. As shown in \tabref{tab:small_batch}, training models with small batch sizes (e.g., 2 or 1) generally increase the error rates. However, such an issue can be addressed by appying AdaptBN to our method. To be more sepcific, we achieve an absolute improvement of 17.9\% and 2.2\% from Source and AdaptBN, respectively, in the batch size of 1.

\paragraph{Number of the source samples for meta networks.} Like previous TTA works~\cite{swr,ttt++, cafa, lim2023ttn} including EATA~\cite{eata}, our approach requires access to the source data for pre-training our proposed meta networks before model deployment. 
In order to cope with the situation where we can only make use of a subset of the source dataset, we study the TTA performance of our method according to the number of accessible source samples. The results are specified in \tabref{tab:num_source} where we use WideResNet-40. 
We observe that our method outperforms the baseline model even with small number of training samples (\eg, 10\% or 20\%) while showing comparable performance with excessively small numbers (\eg 5\%). Note that we still reduce the memory usage of about 51\% compared to EATA.

%% file: table/ablation_arch.tex
\begin{table*}[t]
\vspace{-2.em}
\centering
\subfloat[
\textbf{Visualization of networks variants}
\label{tab:vi_auxnet}
]{
\centering
\begin{minipage}{0.38\linewidth}{\begin{center}

\includegraphics[width=0.98\linewidth]{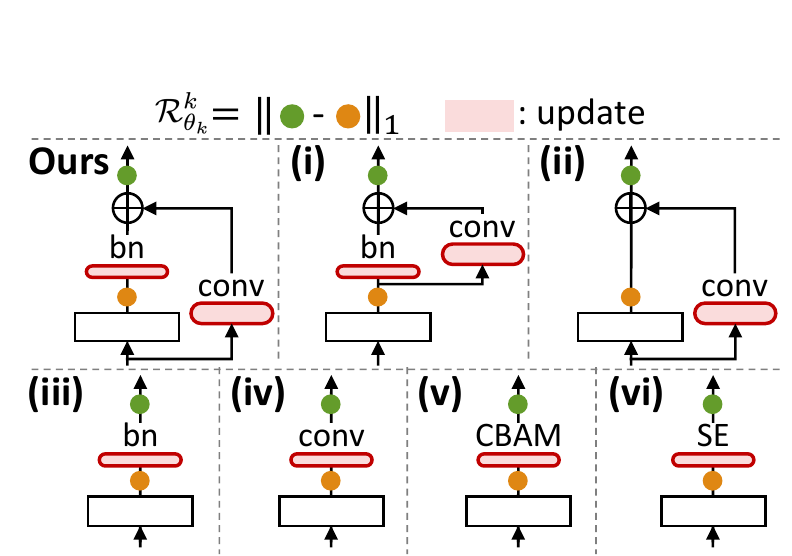}

\end{center}

}\end{minipage}
}
\subfloat[
\textbf{Meta network design (K=5)}
\label{tab:auxnet_ablation}
]{
\begin{minipage}{0.33\linewidth}{\begin{center}

{\renewcommand{\arraystretch}{1.3}
\resizebox{\textwidth}{!}{
\huge
\begin{tabular}{lccc} 
\toprule
\begin{tabular}[c]{@{}l@{}} Avr. err \\\textbf{Arch}\end{tabular}                                     & \begin{tabular}[c]{@{}c@{}}CIFAR10-C\\\textbf{WRN-28}\end{tabular} & \begin{tabular}[c]{@{}c@{}}CIFAR10-C\\\textbf{WRN-40}\end{tabular} & \begin{tabular}[c]{@{}c@{}}CIFAR100-C\\\textbf{WRN-40}\end{tabular}  \\ 
\drule
(i)            & \underbar{18.1}   & \underbar{12.6}    & \underbar{37.2}                                                       \\
(ii) Ours w$\setminus$o BN  & 18.7    & 13.7                                                     & 38.2                                                       \\
(iii) Ours w$\setminus$o Conv    & 20.7   & 14.9                                                     & 40.1                                                       \\
(iv) Conv  & 60.6   & 73.3                                                     & 77.2                                                       \\
(v) CBAM~\cite{cbam}    & 21.4  & 15.1                                                     & 40.9                                                       \\
(vi) SE~\cite{senet}     & 22.3  & 16.2                                                     & 40.5                                                       \\
\rowcolor[rgb]{0.9,0.9,0.9} \textbf{Ours} & \textbf{16.8}  & \textbf{12.1} & \textbf{36.3}                                                      \\
\toprule
\end{tabular}}}
\vspace{+.4em}

\end{center}}

\end{minipage}
}
\subfloat[
\textbf{\#\,of blocks of each partition\,(K=4)}
\label{tab:model_partition}
]{
\begin{minipage}{0.24\linewidth}{\begin{center}

{\renewcommand{\arraystretch}{1.1}
\resizebox{0.92\textwidth}{!}{
{\Large
\begin{tabular}{ccc} 
\toprule
\textbf{Model}         & \textbf{\#Block}         & \textbf{Avg. err}                                  \\ 
\drule
\multirow{3}{*}{{\renewcommand{\arraystretch}{.8} \begin{tabular}[c]{@{}c@{}}WRN-28 (12)\\{\large CIFAR10-C}\end{tabular}}}  & \:\: 3,3,3,3 \:\:                                     & \underbar{17.3}                                      \\
                                                                                & 4,4,2,2                                     & 17.9                                      \\ 
                                                                                & {\cellcolor[rgb]{0.9,0.9,0.9}}2,2,4,4 & {\cellcolor[rgb]{0.9,0.9,0.9}}\textbf{16.9}  \\
\midrule
\multirow{3}{*}{{\renewcommand{\arraystretch}{.8} \begin{tabular}[c]{@{}c@{}}WRN-40 (18)\\{\large CIFAR10-C}\end{tabular}}}  & 4,4,5,5                                     & \underbar{12.8}                                      \\
                                                                                & 6,6,3,3                                     & 13.7                                      \\ 
                                                                                & {\cellcolor[rgb]{0.9,0.9,0.9}}3,3,6,6 & {\cellcolor[rgb]{0.9,0.9,0.9}}\textbf{12.2}  \\
\midrule
\multirow{3}{*}{{\renewcommand{\arraystretch}{.8} \begin{tabular}[c]{@{}c@{}}WRN-40 (18)\\{\large CIFAR100-C}\end{tabular}}} & 4,4,5,5                                     & \underbar{36.9}                                      \\
                                                                                & 6,6,3,3                                     & 38.5                                      \\
                                                                                & {\cellcolor[rgb]{0.9,0.9,0.9}}3,3,6,6 & {\cellcolor[rgb]{0.9,0.9,0.9}}\textbf{36.4}  \\
\toprule
\end{tabular}
}}}
\vspace{+.4em}

\end{center}}

\end{minipage}
}

\vspace{-1.0em}
\caption{\textbf{Architecture ablation experiments.}
\textbf{(a,b)} We compare continual TTA performance on several memory-efficient designs. WRN refers to WideResNet~\cite{wrn} backbone.
\textbf{(c)} We report the performance based on different designs of partitioning the model. The value next to the backbone's name denotes the total number of residual blocks of a model.}
\label{tab:arch_ablations}
\vspace{-1.6em}
\end{table*}

%% file: table/ablation_k_memory.tex
\begin{figure}[t]\centering
\vspace{0.2em}
\includegraphics[width=1\linewidth]{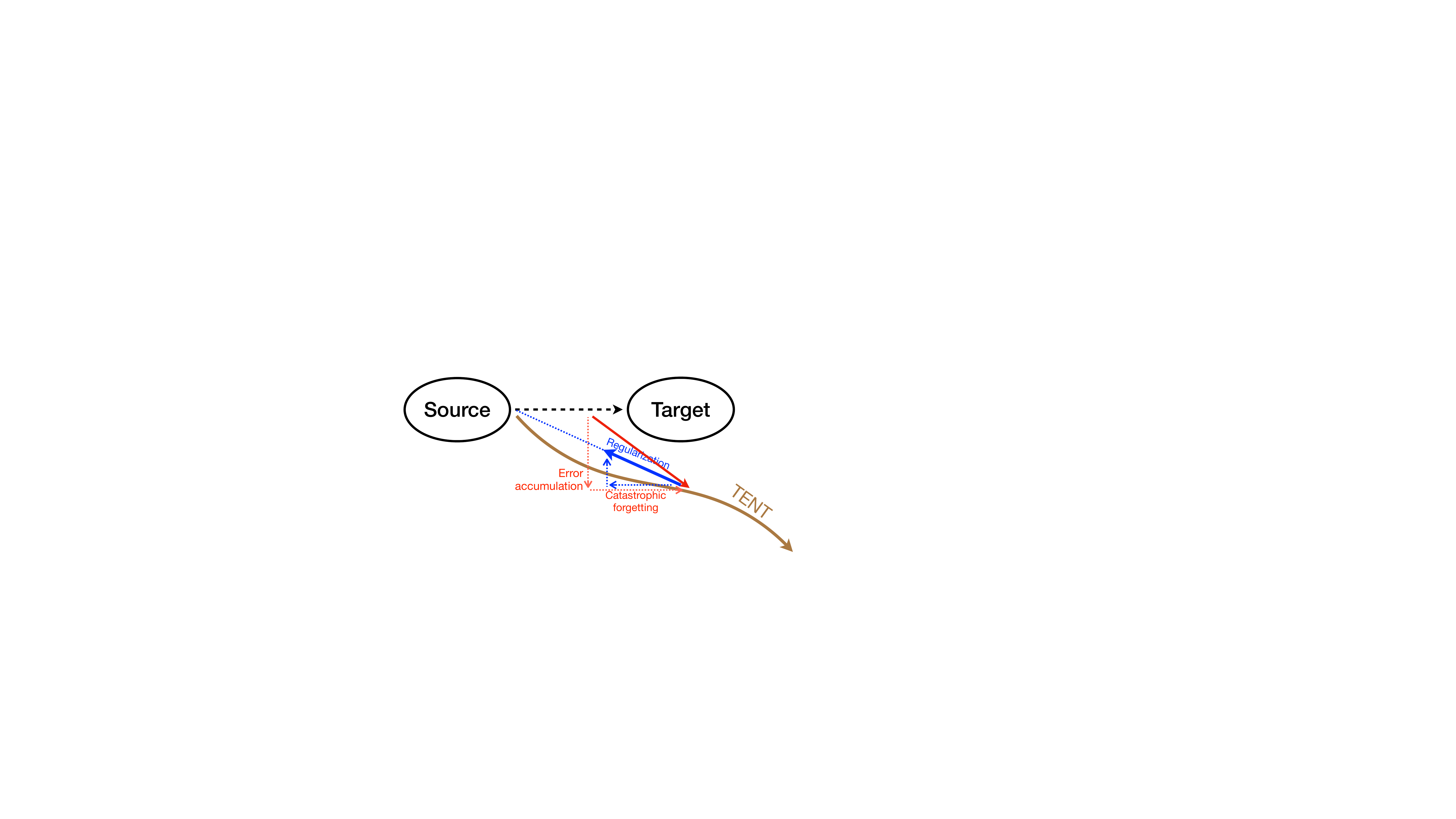}
\vspace{-2.em}
\caption{\textbf{Ablation study of K.} We uniformly divide the encoder of the pre-trained model into the model partition factor K. The x-axis indicates the memory size including both model parameter size and activation size while the y-axis indicates the average error rate. The values in parentheses show the rate of increase for the model parameters compared to the original model.}
\label{fig:k_memory}
\vspace{-1.4em}
\end{figure}

%% file: table/ablation_regular.tex
\begin{figure*}[t]
\centering
\subfloat[
\textbf{Catastrophic forgetting effect}
\label{fig:forgetting}
]{
\centering
\begin{minipage}{0.50\linewidth}{\begin{center}

\includegraphics[width=0.99\linewidth]{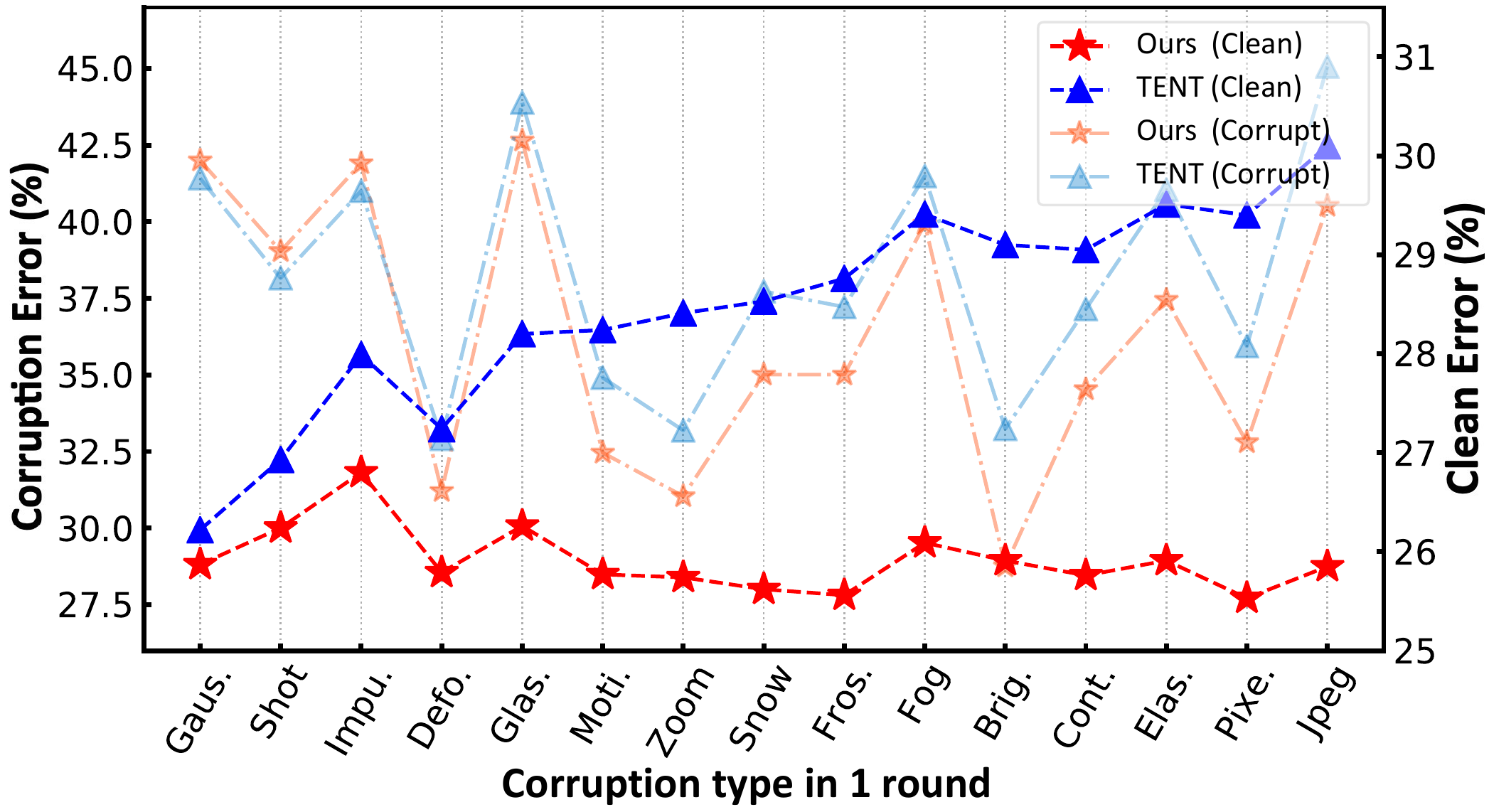}

\end{center}

}\end{minipage}
\vspace{+.15em}
}
\subfloat[
\textbf{Error accumulation effect}
\label{fig:long}
]{
\centering
\begin{minipage}{0.495\linewidth}{\begin{center}

\includegraphics[width=0.95\linewidth]{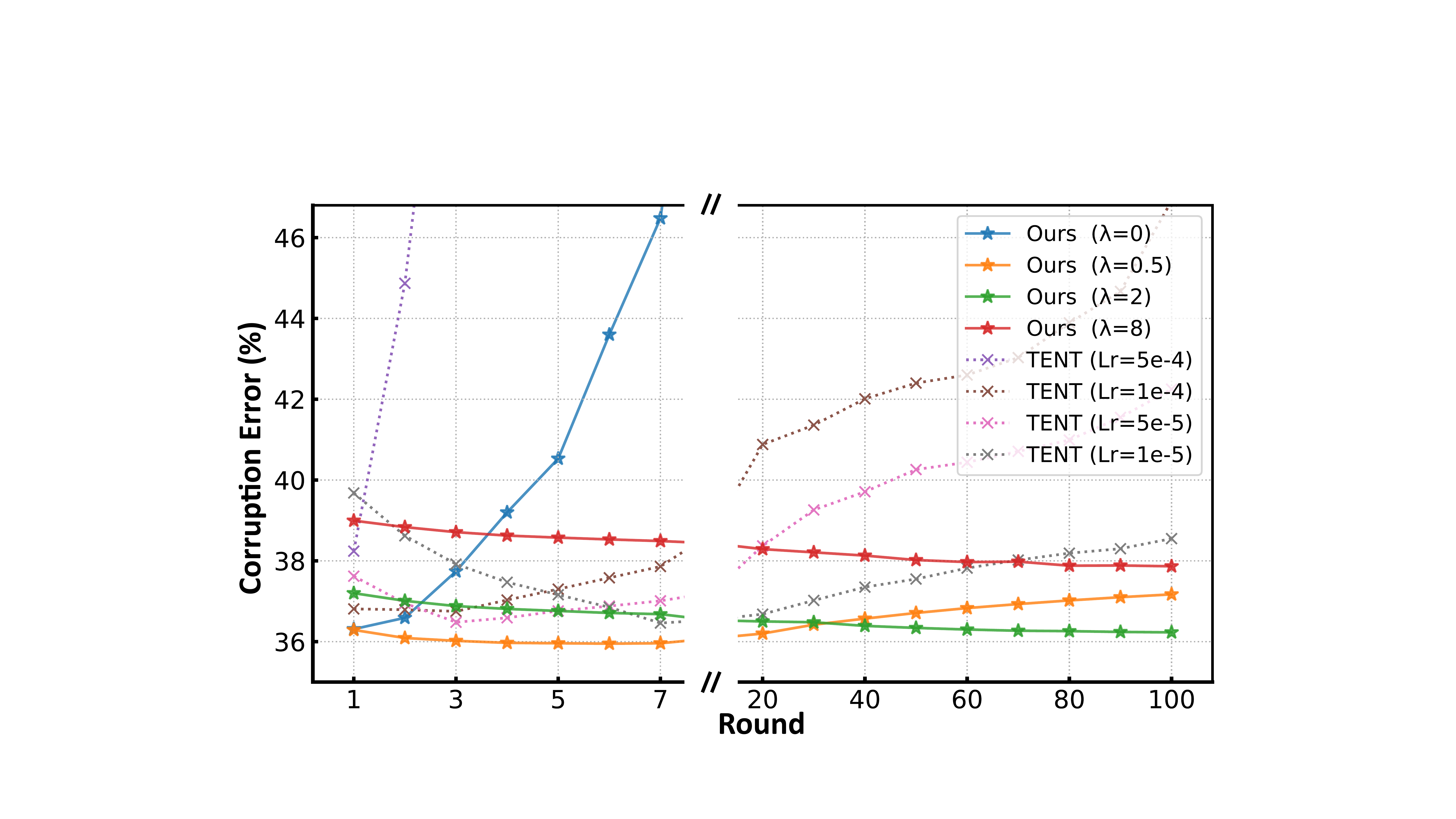}

\end{center}

}\end{minipage}
\vspace{+.52em}
}
\vspace{-.5em}
\caption{\textbf{Regularization ablation experiments}. We conduct experiments with WideResNet-40 on CIFAR100-C. \textbf{(a)} We utilize a test set of the CIFAR-100 dataset to measure clean error after adapting to each corruption. 
Maintaining clean errors at a stable level indicates that our approach helps the model robust to catastrophic forgetting. 
\textbf{(b)} We simulate a long-term adaptation scenario by repeating 100 rounds of 15 corruption sequences. In the absence of regularization, error accumulation can lead to overfitting (\ie, the case of the error increases exponentially). However, our approach does not suffer from such an error accumulation. We set K to 5 in the above experiments.
}
\label{fig:regular_ablations} 
\vspace{-1.8em}
\end{figure*}

%% file: table/ablation_smallbatch.tex
{\renewcommand{\arraystretch}{1.05}
\begin{table}[t]
\centering
\vspace{0.15cm}
\resizebox{.47\textwidth}{!}{
\small
\begin{tabular}{clccccc} 
\toprule
                                                                       & \textbf{Batch size}                                       & \textbf{16}                              & \textbf{8}                               & \textbf{4}                               & \textbf{2}                               & \textbf{1}                                \\
\drule
\multirow{3}{*}{\begin{tabular}[c]{@{}c@{}}Non\\training\end{tabular}} & Source                                                    & 69.7                                     & 69.7                                     & 69.7                                     & 69.7                                     & 69.7                                      \\
                                                                       & BN Stats Adapt~\cite{tbn}                                           & 41.1                                     & 50.2                                     & 59.9                                     & 81.0                                     & 99.1                                      \\
                                                                       & AdaptBN~\cite{adapt_abn}                                                  & 39.1                                     & 41.2                                     & 45.2                                     & 49.0                                     & 54.0                                      \\ 
\hline
\multirow{4}{*}{Training}                                              & Con.~TENT~\cite{tent}                                                & 40.9                                     & 47.8                                     & 58.6                                     & 82.2                                     & 99.0                                      \\
                                                                       & Con.~TENT+AdaptBN                                         & 38.2                                     & 40.2                                     & 43.2                                     & 47.7                                     & 52.2                                      \\
                                                                       & {\cellcolor[rgb]{0.9,0.9,0.9}}\textbf{Ours~{\small (K=5)}}         & {\cellcolor[rgb]{0.9,0.9,0.9}}40.0 & {\cellcolor[rgb]{0.9,0.9,0.9}}45.8 & {\cellcolor[rgb]{0.9,0.9,0.9}}63.4 & {\cellcolor[rgb]{0.9,0.9,0.9}}80.8 & {\cellcolor[rgb]{0.9,0.9,0.9}}99.0  \\
                                                                       & {\cellcolor[rgb]{0.9,0.9,0.9}}\textbf{Ours~{\small (K=5)}+AdaptBN} & {\cellcolor[rgb]{0.9,0.9,0.9}}\textbf{36.9} & {\cellcolor[rgb]{0.9,0.9,0.9}}\textbf{39.3} & {\cellcolor[rgb]{0.9,0.9,0.9}}\textbf{42.2} & {\cellcolor[rgb]{0.9,0.9,0.9}}\textbf{46.5} & {\cellcolor[rgb]{0.9,0.9,0.9}}\textbf{51.8}  \\
\toprule
\end{tabular}}
\small
\vspace{-0.2cm}
\caption{\textbf{Experiments with small batch sizes.} We evaluate all baselines with WideResNet-40 on CIFAR100-C. Con.~TENT is the abbreviation for continual TENT.}
\label{tab:small_batch} 
\vspace{-2.0em}
\end{table}}

%% file: table/main_segment.tex
{\renewcommand{\arraystretch}{1.1}
\begin{table*}[t]

\resizebox{.99\textwidth}{!}{
\Large
\begin{tabular}{l|c|cccc|cccc|cccc|cccc|c}\toprule
Time & &  \multicolumn{16}{l|}{$t\xrightarrow{\hspace*{23cm}}$}  \\ \hline
Round &     & \multicolumn{1}{l}{1} & \multicolumn{1}{l}{} & \multicolumn{1}{l}{} & \multicolumn{1}{l|}{} & \multicolumn{1}{l}{4} & \multicolumn{1}{l}{} & \multicolumn{1}{l}{} & \multicolumn{1}{l|}{} & \multicolumn{1}{l}{7} & \multicolumn{1}{l}{} & \multicolumn{1}{l}{} & \multicolumn{1}{l|}{} & \multicolumn{1}{l}{10} & \multicolumn{1}{l}{} & \multicolumn{1}{l}{} & \multicolumn{1}{l|}{} & \multicolumn{1}{c}{All}  \\ \hline
Method         &  \textbf{Mem. (MB)} & \textbf{Brig.} & \textbf{Fog} & \textbf{Fros.} & \textbf{Snow} & \textbf{Brig.} & \textbf{Fog} & \textbf{Fros.} & \textbf{Snow} &  \textbf{Brig.} & \textbf{Fog} & \textbf{Fros.} & \textbf{Snow} &  \textbf{Brig.} & \textbf{Fog} & \textbf{Fros.} & \textbf{Snow} & \textbf{Mean}             \\ \drule
Source    &    280  & 60.4 & 54.3 & 30.0 & 4.1  & 60.4 & 54.3 & 30.0 & 4.1  & 60.4 & 54.3 & 30.0 & 4.1  & 60.4 & 54.3 & 30.0 & 4.1  & 37.2              \\
BN Stats Adapt~\cite{tbn}   &  280  & 69.1 & 61.0 & 44.8 & 39.1 & 69.1 & 61.0 & 44.8 & 39.1 & 69.1 & 61.0 & 44.8 & 39.1 & 69.1 & 61.0 & 44.8 & 39.1 & 53.6  \\
Continual TENT~\cite{tent}     & 2721        & 70.1 & 62.1 & 46.1 & 40.2 & 62.2 & 53.7 & 44.4 & 37.9 & 50.0 & 41.5 & 31.6 & 26.6 & 39.2 & 32.6 & 25.3 & 22.4 & 42.9 \\ 
\rowcolor[rgb]{0.9,0.9,0.9} \textbf{Ours (K=4)} & \textbf{918}~{\normalsize (66\%}{\scriptsize $\downarrow$}{\small )} & \textbf{70.2} & \textbf{62.4} & \textbf{46.3} & \textbf{41.9} & \textbf{70.0} & \textbf{62.8} & \textbf{46.5} & \textbf{42.2} & \textbf{70.0} & \textbf{62.8} & \textbf{46.5} & \textbf{42.1} & \textbf{70.1} & \textbf{62.8} & \textbf{46.6} & \textbf{42.2} & \textbf{55.3}    
\\ \toprule
\end{tabular}}
\centering
\small
\vspace{-.8em}
\caption{\textbf{Semantic segmentation results in continual test-time adaptation tasks.} We conduct experiments on Cityscapes~\cite{cityscapes} with four weather corruptions~\cite{imagenetC} applied. The four conditions are repeated ten times to simulate continual domain shifts. All results are evaluated based on DeepLabV3Plus-ResNet-50. }
\label{tab:cityscapes}
\vspace{-1.4em}
\end{table*}}

%% file: section/experiment2.tex
We investigate our approach in semantic segmentation. First, we create Cityscapes-C by applying the weather corruptions (brightness, fog, frost, and snow~\cite{imagenetC}) to the validation set of Cityscapes~\cite{cityscapes}. Then, to simulate continual distribution shifts, we repeat the four types of Cityscapes-C ten times. 
In this scenario, we conduct continual TTA using the publicly-available ResNet-50-based DeepLabV3\texttt{+}~\cite{deeplabv3}, which is pre-trained on Cityscapes for domain generalization task~\cite{robusetnet} in semantic segmentation. For TTA, we use the batch size of 2. More details are specified in Appendix C.

\input{table/ablation_num_s}

\paragraph{Results.} 
We report the results based on mean intersection over union (mIoU) in \tabref{tab:cityscapes}. It demonstrates that our approach helps to both minimize memory consumption and performs long-term adaptation stably for semantic segmentation. Unlike continual TENT, our method avoids catastrophic forgetting and error accumulation, allowing us to achieve the highest mIoU score while using 66\% less memory usage in a continual TTA setup. Additional experiment results can be found in Appendix B.

%% file: table/ablation_num_s.tex
{\renewcommand{\arraystretch}{1.1}
\begin{table}[t]
\centering
\resizebox{.47\textwidth}{!}{
\begin{tabular}{c|c|cccc} 
\toprule
         &  \multirow{2}{*}{\begin{tabular}[c]{@{}c@{}}\textbf{EATA~\cite{eata}}~ \\(188MB)\end{tabular}}     &  {\cellcolor[rgb]{0.9,0.9,0.9}}   & \multicolumn{3}{c}{\textbf{\# of source samples}}  \\
Target domain  &  & \multirow{-2}{*}{{\cellcolor[rgb]{0.9,0.9,0.9}}\begin{tabular}[c]{@{}>{\cellcolor[rgb]{0.9,0.9,0.9}}c@{}}\textbf{Ours {\small (K=5)}} \\(92MB)\end{tabular}}  & {\small 10k~(20\%)} & {\small 5k~(10\%)}  & {\small 2.5k~(5\%)}       \\ 
\drule
CIFAR10-C  & 13.0  & {\cellcolor[rgb]{0.9,0.9,0.9}} \textbf{12.1} &  12.4       & 12.9      & 13.1             \\
CIFAR100-C & 37.1  & {\cellcolor[rgb]{0.9,0.9,0.9}} \textbf{36.3} &   36.4       & 36.6      & 37.2             \\
\toprule
\end{tabular}}
\vspace{-0.8em}
\caption{\textbf{Ablation of \# of source samples to warm up the meta networks.} Before deployment, we pre-trained the meta networks using only a subset of the source dataset (\eg, 20\%, 10\%, and 5\%). The memory usage (MB) of each method is also presented.}
\label{tab:num_source}
\vspace{-1.6em}
\end{table}}

%% file: section/conclusion.tex
\vspace{.1em}
This paper proposed a simple yet effective approach that improves continual TTA performance and saves a significant amount of memory, which can be applied to edge devices with limited memory.
First, we presented a memory-efficient architecture that consists of original networks and meta networks. 
This architecture requires much less memory size than the previous TTA methods by decreasing the intermediate activations used for gradient calculations.
Second, in order to preserve the source knowledge and prevent error accumulation during long-term adaptation with noisy unsupervised loss, we proposed self-distilled regularization that controls the output of meta networks not to deviate significantly from the output of the original networks.
With extensive experiments on diverse datasets and backbone networks, we verified the memory efficiency and TTA performance of our approach.
In this regard, we hope that our efforts will facilitate a variety of studies that make test-time adaptation for edge devices feasible in practice.

%% file: section/appendix.tex
\section*{Appendix}
\noindent In this supplementary material, we provide,
\vspace{-.5em}
\begin{enumerate}[label=\Alph*.]
    \setlength\itemsep{-.05em}
    \item Efficiency for TTA methods 
    \item Discussion and further experiments
    \item Further implementation details
    \item Additional ablations 
    \item Baseline details
    \item Results of all corruptions
\end{enumerate}

\section{Efficiency for TTA methods}

\begin{figure}[h]\centering
\includegraphics[width=1\linewidth]{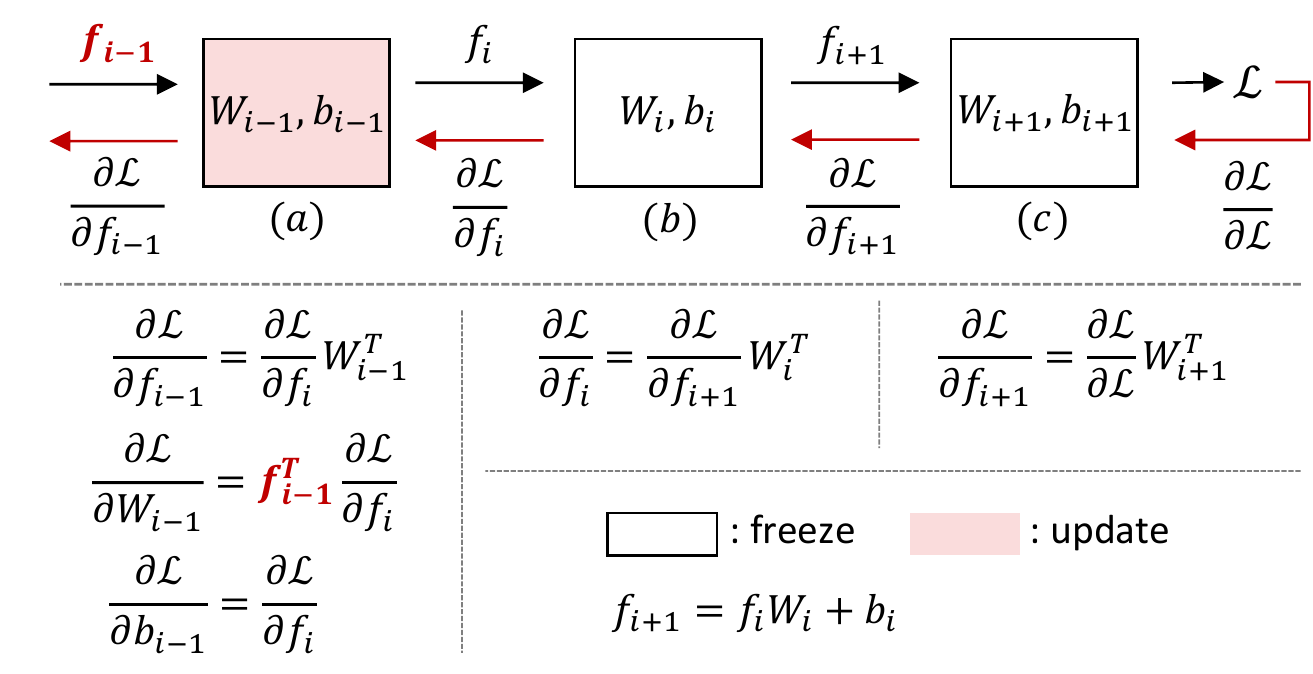}
\vspace{-1.em}
\caption{\textbf{Forward and backward propagation.} The black and red lines refer to forward and backward propagation, respectively. $f$ and ($a,b,c$) are the activations and the linear layers, respectively.}
\label{fig:flow}
\vspace{-.1em}
\end{figure}

\paragraph{Memory efficiency.} 
Existing TTA works~\cite{tent,eata,cotta} update model parameters to adapt to the target domain. 
This process inevitably requires additional memory to store the activations. \figref{fig:flow} describes \equref{eq:backward} of the main paper in more detail. 
For instance,  1) the backward propagation from the layer\,($c$) to the layer\,($b$) can be accomplished without saving intermediate activations {\small $f_{i}$ and $f_{i+1}$}, since it only requires {\small $\frac{\partial \loss}{\partial f_{i+1}}{=}\frac{\partial \loss}{\partial \loss}\mathcal{W}_{i+1}^{T}$ and
$\frac{\partial \loss}{\partial f_{i}}{=}\frac{\partial \loss}{\partial f_{i+1}}\mathcal{W}_{i}^{T}{=}\frac{\partial \loss}{\partial \loss} \mathcal{W}_{i+1}^{T}\mathcal{W}_{i}^{T}$} operations. 2) During the forward propagation, the learnable layer\,($a$) has to store the intermediate activation {\small $\textcolor{darkred}{f_{i-1}}$} to calculate the weight gradient {\small 
 $\frac{\partial \loss}{\partial \mathcal{W}_{i-1}}{=}\textcolor{darkred}{f_{i-1}^{T}}\frac{\partial \loss}{\partial f_{i}}$}.

\paragraph{Computational efficiency.} 
Wall-clock time and floating point operations (FLOPs) are standard measures of computational cost. We utilize wall-clock time to compare the computational cost of TTA methods since most libraries computing FLOPs only support inference, not training.

\input{table/time}

Unfortunately, wall-clock time of EATA~\cite{eata} and our approach can not truly represent its computational efficiency since the current Pytorch version~\cite{pytorch} does not support fine-grained implementation~\cite{tinytl}.
For example, EATA filters samples to improve its computational efficiency. However, its gradient computation is performed on the full mini-batch, so the wall-clock time for backpropagation in EATA is almost the same as that of TENT~\cite{tent}. In our approach, our implementation follows Algorithm~\ref{alg:ecotta} to make each regularization loss $\mathcal{R}_{\theta_k}^{k}$ applied to parameters of k-th group of meta networks $\theta_k$ in \equref{eq:totalloss}. In order to circumvent such an issue, the authors of EATA report the theoretical time, which assumes that PyTorch handles gradient backpropagation at an instance level. Similar to EATA, we also report both theoretical time and wall-clock time in \tabref{tab:time}. To compute the theoretical time of our approach, we simply subtract the time for re-forward (in Algorithm~\ref{alg:ecotta}) from wall-clock time. We emphasize that this is mainly an engineering-based issue, and the optimized implementation can further improve computational efficiency.~\cite{eata}.

Using a single NVIDIA 2080Ti GPU, we measure the total time required to adapt to all 15 corruptions, including the time to load test data and perform TTA.
The results in \tabref{tab:time} show that our proposed method requires negligible overhead compared to CoTTA~\cite{cotta}. For example, CoTTA needs approximately 10 times more training time than Continual TENT~\cite{tent} with WideResNet-40. 
Note that meta networks enable our approach to use 80\% and 58\% less memory than CoTTA and EATA, even with such minor extra operations.

\input{table/pseudocode}

\section{Discussion and further experiments}
\label{sec:discuss}

\paragraph{Comparison on gradually changing setup.} In \tabref{tab:main_cifar} and \tabref{tab:main_imagenet}, 
we evaluate all methods on the continual TTA task, proposed in CoTTA~\cite{cotta} and EATA~\cite{eata}, where we continually adapt the deployed model to each corruption type sequentially. Additionally, we conduct experiments on the gradually changing setup. This gradual setup, proposed in CoTTA, represents the sequence by gradually changing severity for the 15 corruption types:
\begingroup
\footnotesize
${\underbrace{{\dots}2{\xrightarrow{}}1}_{\text{t-1 and before}}}{\xrightarrow[type]{\small change}}{\underbrace{1{\xrightarrow{}}2{\xrightarrow{}}3{\xrightarrow{}}4{\xrightarrow{}}5{\xrightarrow{}}4{\xrightarrow{}}3{\xrightarrow{}}2{\xrightarrow{}}1}_{\text{corruption type t, gradually changing severity}}}{\xrightarrow[type]{\small change}}\underbrace{1{\xrightarrow{}}2{\dots}}_{\text{t+1 and on}}$ ,
\endgroup
\vspace{.3em}

\noindent
The results in \tabref{tab:gradualsetup} indicate that our approach outperforms previous TTA methods~\cite{tent,eata,cotta} even with the gradually changing setup. 

\paragraph{Comparisons with methods for parameter efficient transfer learning.} While our framework may be similar to parameter-efficient transfer learning (PETL)~\cite{hu2022lora,houlsby2019parameter,sun2019meta} in that only partial parameters are updated during training time for PETL or test time for TTA, we utilized meta networks to minimize intermediate activations, which is crucial for memory-constrained edge devices. 
We conduct experiments by applying a PETL method~\cite{houlsby2019parameter} to the TTA setup. 
The adapter module is constructed by using 3x3 Conv and ReLU layers as the projection layer and the nonlinearity, respectively, and these modules are attached after each residual block of the backbone networks.
The \tabref{tab:petl} shows that PETA+SDR needs a 177\% increase in memory usage with a 6.1\% drop in performance, compared to our method.

{\renewcommand{\arraystretch}{1.1}
\begin{table}[t]
\centering
\resizebox{.48\textwidth}{!}{
{\large
\begin{tabular}{lcccc} 
\toprule
\textbf{Method} & Con. TENT~\cite{tent} & EATA~\cite{eata} & CoTTA~\cite{cotta} & {\cellcolor[rgb]{0.9,0.9,0.9}}~~Ours (K=4)~~  \\ 
\drule
Avg.\,err\,(\%)     & 38.5 & 31.8 & 32.5 & {\cellcolor[rgb]{0.9,0.9,0.9}}\textbf{31.4}     \\ 
Mem.\,(MB)    & 188 & 188 & 409 & {\cellcolor[rgb]{0.9,0.9,0.9}}\textbf{80}\,{\small (58, 80\%}{\scriptsize $\downarrow$}{\small )} \\
\toprule
\end{tabular}}}
\vspace{-.5em}
\caption{\textbf{Comparision on gradually changing setup.} To conduct experiments, we use WRN-40 backbone on CIFAR100-C. The values in parentheses refer to memory reduction rates compared to
TENT/EATA and CoTTA, sequentially.}
\label{tab:gradualsetup}
\vspace{-.2em}
\end{table}}

{\renewcommand{\arraystretch}{1.1}
\begin{table}[t!]
\centering
\resizebox{.48\textwidth}{!}{
{\large
\begin{tabular}{l|c|ccc} 
\toprule
\textbf{Method} & Con. TENT~\cite{tent} & ~~PETL~\cite{houlsby2019parameter}~~ & PETL+SDR & {\cellcolor[rgb]{0.9,0.9,0.9}}~~~Ours (K=4)~~~  \\ 
\drule
Avg.\,err\,(\%)     & 38.3 & 73.3 & 42.5 & {\cellcolor[rgb]{0.9,0.9,0.9}}\textbf{36.4}     \\ 
Mem.\,(MB)          & 188  & 141  & 141  & {\cellcolor[rgb]{0.9,0.9,0.9}}\textbf{80} \\
\toprule
\end{tabular}}}
\vspace{-.5em}
\caption{\textbf{Comparisons with methods for PETL.} We compare our method with methods~\cite{houlsby2019parameter} for parameter-efficient transfer learning (PETL) with WRN-40 on CIFAR100-C. PETL+SDR refers to PETL with our proposed self-distilled regularization.}
\label{tab:petl}
\vspace{-.2em}
\end{table}}

{\renewcommand{\arraystretch}{1.1}
\begin{table}[t]
\centering
\resizebox{.45\textwidth}{!}{
{\large
\begin{tabular}{c|c|ccccc} 
\toprule
\textbf{~Round~} & Con. TENT & ~~TS~~ & ~~DO~~ & ~~LS~~ & ~~KD~~ & {\cellcolor[rgb]{0.9,0.9,0.9}} ~~Ours (K=4)~\\ 
\drule
1    & 38.3  & 37.4 & 41.0 & 38.4 & 39.8 & {\cellcolor[rgb]{0.9,0.9,0.9}}\textbf{36.4}     \\ 
10   & 99.0  & 96.1 & 96.3 & 41.1 & 40.4 & {\cellcolor[rgb]{0.9,0.9,0.9}}\textbf{36.3}     \\
\toprule
\end{tabular}}}
\vspace{-.5em}
\caption{\textbf{Comparisons with methods for continual learning.} We report an average error rate (\%) of 15 corruptions using WRN-40 on CIFAR100-C. In the table, TS: Entropy minimization with temperature scaling~\cite{guo2017calibration}, DO: Dropout~\cite{dropout}, LS: Label smoothing with the pseudo label~\cite{labelsmooth}, and KD: Knowledge distillation~\cite{kd}.}
\label{tab:cl}
\vspace{-1.em}
\end{table}}

\paragraph{Comparisons with methods for continual learning.}
Typical continual learning (CL) and continual TTA assume supervised and unsupervised learning, respectively. However, since both are focused on alleviating catastrophic forgetting, we believe that CL methods can also be applied in continual TTA settings.
The methods for addressing catastrophic forgetting can be divided into regularization- and replay-based methods. 
The former can be subdivided into weight regularization (\eg, CoTTA~\cite{cotta} and EATA~\cite{eata}) and knowledge distillation~\cite{kd}, while the latter includes GEM~\cite{lopez2017gradient} and dataset distillation~\cite{deng2022remember}.
Suppose dataset distillation is applied to the continual TTA setup; for example, we can periodically replay synthetic samples distilled from the source dataset to prevent the model from forgetting the source knowledge during TTA.
Notably, our self-distilled regularization (SDR) is superior to conventional CL methods in terms of the efficiency of TTA in on-device settings. Specifically, unlike previous regularization- or replay-based methods, we do not require storing a copy of the original model or a replay-and-train process.

To further compare our SDR with existing regularization methods, we conduct experiments while keeping our architecture and adaptation loss but replacing SDR with other regularizations, as shown in \tabref{tab:cl}. The results demonstrate that our SDR achieves superior performance compared to other regularizations. In addition, Knowledge distillation~\cite{kd} alleviates the error accumulation effect in long-term adaptation (\eg, round\,10), while showing limited performance for adapting to the target domain.

\paragraph{Superiority of our approach compared to existing TTA methods.} Our work focuses on proposing an efficient architecture for continual TTA, which has been overlooked in previous TTA studies~\cite{tent,cotta,chen2022contrastive,ttt++,swr,lim2023ttn} by introducing meta networks and self-distilled regularization, rather than adaptation loss such as entropy minimization proposed in TENT~\cite{tent} and EATA~\cite{eata}. Thus, our method can be used with various adaptation losses. Moreover, even though our self-distilled regularization can be regarded as a teacher-student distillation from original networks to meta networks, it does not require a large activation size or the storage of an extra source model, unlike CoTTA~\cite{cotta}.

{\renewcommand{\arraystretch}{1.05}
\begin{table}[t]
\centering
\resizebox{.48\textwidth}{!}{
{\large
\begin{tabular}{lccccc} 
\toprule
\textbf{Method}                                   & \textbf{Mem. (MB)}          & \textbf{Round\,1} & \textbf{Round\,4} & \textbf{Round\,7} & \textbf{Round\,10}  \\ 
\drule
Source                                   & 280                & 37.2    & 37.2    & 37.2    & 37.2      \\
Con. TENT                                & 2721                & 54.6    & 49.6    & 37.4    & 29.9      \\
Con. TENT~\textsuperscript{\textbf{*}}                              & 2721                & 56.5    & 52.7    & 42.7    & 36.5      \\
CoTTA~\textsuperscript{\textbf{*}}                                    & 6418               & \textbf{56.7}    & 56.7    & 56.7    & 56.7      \\
\rowcolor[rgb]{0.9,0.9,0.9} Ours   & \textbf{918} {\small (66, 85\%}{\scriptsize $\downarrow$}{\small )} & 55.2    & 55.4    & 55.4    & 55.4      \\
\rowcolor[rgb]{0.9,0.9,0.9} Ours~\textsuperscript{\textbf{*}} & \textbf{918} {\small (66, 85\%}{\scriptsize $\downarrow$}{\small )} & \textbf{56.7}    & \textbf{56.8}    & \textbf{56.9}    & \textbf{56.9}      \\
\toprule
\end{tabular}}}
\vspace{-.5em}
\caption{\textbf{Further experiments in semantic segmentation.} We represent the results based on mean intersection over union (mIoU). \textsuperscript{\textbf{*}} means that the method utilizes the same cross-entropy consistency loss. The values in parentheses refer to memory reduction rates compared to TENT/EATA and CoTTA, sequentially.}
\label{tab:add_segment}
\vspace{-.2em}
\end{table}}

\input{table/blocks}

In addition to the results in \tabref{tab:cityscapes}, we improve the segmentation experiments by comparing our approach with CoTTA~\cite{cotta}. As we aforementioned, our approach has scalability with diverse adaptation loss. Thus, as shown in \tabref{tab:add_segment}, we additionally apply cross-entropy consistency loss\textsuperscript{\textbf{*}} with multi-scaling input as proposed in CoTTA, where we use the multi-scale factors of [0.5, 1.0, 1.5, 2.0] and flip. Our method not only achieves comparable performance with 85\% less memory than CoTTA, but shows consistent performance even for multiple rounds while continual TENT~\cite{tent} suffers from the error accumulation effect.

\section{Further implementation details}
\label{sec:furtherdetails}
\paragraph{Partition of a pre-trained model.}
As illustrated in \figref{fig:approach}, the given pre-trained model consists of three parts: classifier, encoder, and input conv, where the encoder denotes layer1 to 4 in the case of ResNet.
Our method is applied to the encoder and we divide it into K parts.
\tabref{tab:blocks} describes the details of the number of residual blocks for each part of the encoder. 
Our method is designed to divide the shallow layers more (\ie, densely) than the deep layers, improving the TTA performance as shown in \tabref{tab:model_partition}.

\input{table/kernel_size}
\input{table/transformation}

\input{table/arch_variants}

\paragraph{Convolution layer in meta networks.}
As the hyperparameters of the convolution layer\footnote{\scriptsize \href{https://pytorch.org/docs/stable/generated/torch.nn.Conv2d.html}{https://pytorch.org/docs/stable/generated/torch.nn.Conv2d.html}},
we set the bias to false and 
the stride to two if the corresponding part of the encoder includes the stride of two; otherwise, one.
As shown in the gray area in \tabref{tab:kernel}, we conduct experiments by modifying the kernel size and padding for each architecture.
To be more specific, we obtain better performances by setting the kernel size to three with WideResNet (with 10\% additional number of model parameters).
On the other hand, utilizing the kernel size of three with ResNet leads to significant increases in parameters and memory sizes. Thus, we use one and three as the kernel size with ResNet and WideResNet, respectively.

\paragraph{Warming up meta networks.} Before the model deployment, we warm up meta networks with the source data by applying the following transformations, which prevent the meta networks from being overfitted to the source domain. 

Regardless of the pre-trained model's architecture and pre-training method, we use the same transformations to warm up meta networks. Even for WideResNet-40 pre-trained with AugMix~\cite{augmix}, a strong data augmentation technique, the following simple transformations are enough to warm up the meta networks. In addition, we provide the ablation of the combination of transformations in \tabref{tab:transformation}.
{\scriptsize
\begin{lstlisting}[language=Python]
from torchvision import transforms as T

TRANSFORMS = torch.nn.Sequential(
    RandomApply(T.ColorJitter(0.4,0.4,0.4,0.1), p=0.4)
    RandomApply(T.GaussianBlur((3,3), p=0.2)
    T.RandomGrayscale(P=0.1))
\end{lstlisting}} 

\paragraph{Semantic segmentation.} 
For semantic segmentation experiments, we utilize ResNet-50-based DeepLabV3\texttt{+}~\cite{deeplabv3} from RobustNet repository\footnote{\href{https://github.com/shachoi/RobustNet}{https://github.com/shachoi/RobustNet}}~\cite{robusetnet}. We warm up the meta networks on the train set of Cityscapes~\cite{cityscapes} with SGD optimizer with the learning rate of 5e-2 and the epoch of 5. Image transformations follow the implementation details of \cite{robusetnet}. After model deployment, we perform TTA using SGD optimizer with the learning rate of 1e-5, the image size of 1600$\times$800, the batch size of 2, and the importance of regularization $\lambda$ of 2. The main loss for adaptation is same as $\loss^{ent}$ in \equref{eq:entloss}.

\section{Additional ablations}

\paragraph{Main task loss for adaptation.} To adapt to the target domain effectively, selecting the main task loss for adaptation is a non-trivial problem. So, we conduct a comparative experiment on three types of adaptation loss: $\loss^{1}$) entropy minimization~\cite{entmin}, $\loss^{2}$) entropy minimization with mean entropy maximization~\cite{entmax}, and $\loss^{3}$) filtering samples using entropy minimization~\cite{eata}. With a mini-batch of $N$ test images, the three adaptation losses are formulated as follows:
\begin{flalign}
    &\loss^{1} = \frac{1}{N}\sum_{i=1}^N H(\hat{y}_i), \label{eq:ent_min}\\ 
    &\loss^{2} = \lambda_{m_{1}} \frac{1}{N}\sum_{i=1}^N H(\hat{y}_i) - \lambda_{m_{2}} H(\overline{y}),  \label{eq:ent_max} \\
    &\loss^{3} =  \frac{1}{N}\sum_{i=1}^N \mathbb{I}_{\{H(\hat{y}_i) < H_0\}} 	\cdot H(\hat{y}_i), \label{eq:ent_t}
\end{flalign}\noindent
where $\hat{y}_i$ is the logits output of $i$-th test data, $\overline{y} =\frac{1}{N}\sum_{i=1}^N p(\hat{y}_{i})$, $H(y)=-\sum_C p(y) \log p(y)$, p($\cdot$) is the softmax function, C is the number of classes, and $\mathbb{I}_{\{\cdot\}}$ is an indicator function. $\lambda_{m_{1}}$ and $\lambda_{m_{2}}$ indicate the importance of each term in \equref{eq:ent_max} which are set to 0.2 and 0.25, respectively, following SWR\&NSP~\cite{swr}. The entropy threshold $H_0$ is set to $0.4 \times \ln C$ following EATA~\cite{eata}.

The results are described in \tabref{tab:mainloss}. Particularly, applying any of the three losses, our method achieves comparable performance to EATA. Among them, using $\loss^{3}$ of \equref{eq:ent_t} achieves the lowest error rate in most cases. Therefore, we apply $\loss^{3}$ to our approach as mentioned in \secref{sec:approach1}.

\input{table/main_task_loss}
\input{table/reg_loss}

\paragraph{Components of meta networks.} As shown in \tabref{tab:add_arch_ablations}, we conduct an ablation study on each element of our proposed meta networks. We observe that the affine transformation is more critical than standardization in a BN layer after the original networks. Specifically, removing the standardization (variant II) causes less performance drop than removing the affine transformation (variant I). In addition, using only a conv layer in conv block (variant VI) also cause performance degradation, so it is crucial to use the ReLU and BN layers together in the conv block.

\paragraph{Loss function choice of our regularization.} As mentioned in \secref{sec:approach2}, self-distilled regularization loss computes the mean absolute error (\ie, L1 loss) of \equref{eq:regloss}. 
This loss regularizes the output $\tilde{x}_k$ of each $k$-th group of the meta networks not to deviate from the output $x_k$ of each $k$-th part of frozen original networks.
The mean squared error (\ie, MSE loss) also can be used to get a similar effect which is defined as:
\begin{equation}
\label{eq:mse}
    MSE = (\tilde{x}_k-x_k)^2.
\end{equation}
\noindent
We compare two kinds of loss functions for our regularization in \tabref{tab:regloss}. 
By observing a marginal performance difference, our method is robust to the loss function choice.

\paragraph{Robustness to the importance of regularization $\lambda$.} 
We show that our method is robust to the regularization term $\lambda$. 
We conduct experiments using a wide range of $\lambda$ as shown in \figref{fig:regular_ablations} and the following table.

\input{table/reg_scale}

\noindent
The experiments are performed with WideResNet-40 on CIFAR100-C. 
When $\lambda$ is changed from 0.5 to 1, the performance difference was only 0.27\% in the first round. 
We also test $\lambda$ to be extremely large (\eg, 5, 8, and 10). 
Since setting $\lambda$ to 10 may mean that we hardly adapt the meta networks to the target domain, the error rate (39.58\%) with $\lambda$ of 10 was close to the one (41.1\%) of BN Stats Adapt~\cite{tbn}.

\section{Baseline details}

\subsection{TTA works}
\label{sec:ttaworks}
We refer to the baselines for which the code was officially released: TENT\footnote{\scriptsize \href{https://github.com/DequanWang/tent}{https://github.com/DequanWang/tent}}, TTT++\footnote{\scriptsize \href{https://github.com/vita-epfl/ttt-plus-plus}{https://github.com/vita-epfl/ttt-plus-plus}}, CoTTA\footnote{\scriptsize \href{https://github.com/qinenergy/cotta}{https://github.com/qinenergy/cotta}}, EATA\footnote{\scriptsize \href{https://github.com/mr-eggplant/EATA}{https://github.com/mr-eggplant/EATA}}, and NOTE\footnote{\scriptsize \href{https://github.com/TaesikGong/NOTE}{https://github.com/TaesikGong/NOTE}}. 
We did experiments on their code by adding the needed data loader or pre-trained model loader. 
In this section, implementation details of the baselines are provided.

\paragraph{BN Stats Adapt~\cite{tbn}} is one of the non-training TTA approaches. It can be implemented by setting the model to the train mode\footnote{\scriptsize \href{https://pytorch.org/docs/stable/generated/torch.nn.Module.html\#torch.nn.Module.train}{pytorch.org/docs/stable/generated/torch.nn.Module.html\#torch.nn.Module.train}} of Pytorch~\cite{pytorch} during TTA. 

\paragraph{TTT+++~\cite{ttt++}} was originally implemented as the offline adaptation, \ie, multi-epoch training. So, we modified their setup to continual TTA. We further tuned the learning rate as 0.005 and 0.00025 for adapting to CIFAR10-C and CIFAR100-C, respectively.

\paragraph{NOTE~\cite{note}} proposed the methods named IABN and PBRS with taking account of temporally correlated target data. However, our experiments were conducted with target data that was independent and identically distributed (i.i.d.). Hence, we adapted NOTE-i.i.d (\ie, NOTE* in their git repository), which is a combination of TENT~\cite{tent} and IABN without using PBRS. We fine-tuned the $\alpha$ of their main paper (\ie, self.k in the code\footnote{\scriptsize \href{https://github.com/TaesikGong/NOTE/blob/main/utils/iabn.py}{https://github.com/TaesikGong/NOTE/blob/main/utils/iabn.py}}) to 8 and the learning rate to 1e-5.

\paragraph{Others}~(\eg, TENT~\cite{tent}, SWR\&NSP~\cite{swr}, CoTTA~\cite{cotta}, and EATA~\cite{eata}). We utilized the best hyperparameters specified in their paper and code. In the case where the batch size of their works (\eg, 200 and 256) differs from one for our experiments (\eg, 64), we decreased the learning rate linearly based on the batch size~\cite{goyal2017accurate}.

\paragraph{AdaptBN~\cite{adapt_abn}.} We set the hyperparameter $N$ of their main paper to 8. When AdaptBN is employed alongside TENT or our approach, we set the learning rate to 1e-5 or 5e-6~\cite{lim2023ttn}. 

\subsection{On-device learning works}
To unify the backbone network as ResNet-50~\cite{resnet}, we reproduced the following works by referencing their paper and published code: TinyTL\footnote{\scriptsize \href{https://github.com/mit-han-lab/tinyml/tree/master/tinytl}{https://github.com/mit-han-lab/tinyml/tree/master/tinytl}}, Rep-Net\footnote{\scriptsize \href{https://github.com/ASU-ESIC-FAN-Lab/RepNet}{https://github.com/ASU-ESIC-FAN-Lab/RepNet}}, and AuxAdapt. This section presents additional implementation details for reproducing the above three works. 

\input{table/cifar10_all}
\input{table/cifar100_all}

\paragraph{TinyTL~\cite{tinytl}.}
We attach the LiteResidualModules\footnote{\scriptsize \href{https://github.com/mit-han-lab/tinyml/blob/master/tinytl/tinytl/model/modules.py}{https://github.com/mit-han-lab/tinyml/blob/master/tinytl/tinytl/model/modules.py}} to layer1 to 4 in the case of ResNet-50\footnote{\scriptsize \href{https://github.com/pytorch/vision/blob/main/torchvision/models/resnet.py}{https://github.com/pytorch/vision/blob/main/torchvision/models/resnet.py}}. As the hyperparameters of the LiteResidualModules, the hyperparameter \textit{expand} is set to 4 while the other hyperparameters follow the default values.

\paragraph{Rep-Net~\cite{repnet}.} We divide the encoder of ResNet-50 into six parts, as each part of the encoder has 2,2,3,3,3,3 residual blocks (\eg, BasicBlock or Bottleneck in Pytorch) from the shallow to the deep parts sequentially. Then, we connect the ProgramModules\footnote{\scriptsize \href{https://github.com/ASU-ESIC-FAN-Lab/RepNet/blob/master/repnet/model/reprogram.py}{github.com/ASU-ESIC-FAN-Lab/RepNet/blob/master/repnet/model/reprogram.py}} to each corresponding part of the encoder. For the ProgramModule, we set the hyperparameter \textit{expand} to 4 while the rest hyperparameters are used as their default values. We copy the input conv of ResNet-50 and make use of it as the input conv of Rep-Net.

\paragraph{AuxAdapt~\cite{auxadapt}.} We use ResNet-18 as the AuxNet. We create pseudo labels by fusing the logits output of ResNet-50 and ResNet-18, and optimize all parameters of ResNet-18 using the pseudo labels with cross-entropy loss.

\paragraph{Warming up the additional modules.} Before model deployment, we pre-train the additional modules (\ie, the LiteResidualModule of TinyTL~\cite{tinytl}, the ProgramModule of Rep-Net~\cite{repnet}, and the AuxNet of AuxAdapt~\cite{auxadapt}) on the source data using the same strategy warming up the meta networks as mentioned in \secref{sec:furtherdetails}.

\section{Results of all corruptions}
We report the error rates (\%) of all corruptions on continual TTA and memory consumption (MB) including the model parameters and activations in \tabref{tab:cifar10_all} and \tabref{tab:cifar100_all}. These tables contain additional details to \tabref{tab:main_cifar}.

%% file: table/time.tex
{\renewcommand{\arraystretch}{1.1}
\begin{table}[b!]
\centering
\resizebox{.475\textwidth}{!}{
\begin{tabular}{lcccc} 
\toprule
\multicolumn{5}{c}{{\cellcolor[rgb]{0.85,0.85,0.85}}WideResNet-40~\cite{wrn}}                              \\
     & \textbf{Avg. err} & \textbf{Mem.~(MB)} & \textbf{Theo. time} & \textbf{Wall time}  \\
\drule
Source                                       & 69.7     & 11       & -                & 40s        \\
Con. TENT~\cite{tent}                               & 38.3     & 188      & -                & 2m 18s     \\
CoTTA~\cite{cotta}                                        & 38.1     & 409      & -                & \textcolor{darkred}{22m 52s}    \\
EATA~\cite{eata}                                         & 37.1     & 188      & 2m 8s           & 2m 22s     \\
\rowcolor[rgb]{0.9,0.9,0.9} Ours (K=4) & \underbar{36.4}     & \textbf{80}~{\footnotesize (80, 58\%}{\scriptsize $\downarrow$)}       & 2m 27s           & 2m 49s     \\
\rowcolor[rgb]{0.9,0.9,0.9} Ours (K=5) & \textbf{36.3}     & \underbar{92}~{\footnotesize (77, 51\%}{\scriptsize $\downarrow$)}       & 2m 31s           & 2m 52s     \\ 
\toprule
\multicolumn{5}{c}{{\cellcolor[rgb]{0.85,0.85,0.85}}ResNet-50~\cite{resnet}}                                  \\
  & \textbf{Avg. err} & \textbf{Mem.~(MB)} & \textbf{Theo. time} & \textbf{Wall time}   \\
\drule
Source                                       & 73.8     & 91       & -                & 1m 8s     \\
Con. TENT~\cite{tent}                               & 45.9     & 926      & -                & 4m 2s     \\
CoTTA~\cite{cotta}                                        & 40.2     & 2064     & -                & \textcolor{darkred}{38m 24s} \\
EATA~\cite{eata}                                         & 39.9     & 926      & 3m 45s          & 4m 15s     \\
\rowcolor[rgb]{0.9,0.9,0.9} Ours (K=4) & \underbar{39.5}     & \textbf{296}~{\footnotesize (86, 68\%}{\scriptsize $\downarrow$)}      & 4m 16s           & 4m 41s     \\
\rowcolor[rgb]{0.9,0.9,0.9} Ours (K=5) & \textbf{39.3}     & \underbar{498}~{\footnotesize (76, 46\%}{\scriptsize $\downarrow$)}      & 4m 26s           & 5m 14s     \\
\toprule
\end{tabular}}
\vspace{-.5em}
\caption{\textbf{Comparison of training time on CIFAR100-C.} We report both theoretical time (in short, theo. time) and wall-clock time, taking to adapt to all 15 corruption types. Theoretical time is calculated by assuming that the ML frameworks (\eg, Pytorch~\cite{pytorch}) provide fine-grained implementations~\cite{tinytl}. Con. TENT refers to continual TENT.}
\label{tab:time}
\vspace{-.5em}
\end{table}}

%% file: table/pseudocode.tex
\begin{algorithm}[t!]
   \caption{\small PyTorch-style pseudocode for EcoTTA.}
   \label{alg:ecotta}
    \definecolor{codeblue}{rgb}{0.25,0.5,0.5}
    \lstset{
      basicstyle=\fontsize{7.2pt}{7.2pt}\ttfamily\bfseries,
      commentstyle=\fontsize{7.2pt}{7.2pt}\color{codeblue},
      keywordstyle=\fontsize{7.2pt}{7.2pt},
    }
\begin{lstlisting}[language=python]
# img_t: test image
# model: original and meta networks
#
# ent_min(): Entropy minimization loss
# Detach_parts(): Detach the graph connection 
#                 between each partition of networks
# Attach_parts(): Attach the graph connection
#                 between each partition of networks

for img_t in test_loader:
    # 1. Forward
    output = model(img_t)
    # 2. Compute entropy loss 
    loss_ent = ent_min(output)
    loss_ent.backward()
    
    # 3. Re-forward 
    # (This process is not required
    # in fine-grained ML frameworks.)
    Detach_parts(model)
    _ = model(img_t)   
    
    # 4. Compute regularization loss
    reg_loss = 0
    for k_th_meta in meta_networks:
        reg_loss += k_th_meta.get_l1_loss()
    reg_loss.backward()
    
    # 5. Update params of meta networks
    optim.step()
    optim.zero_grad()
    
    Attach_parts(model)
\end{lstlisting}
\vspace{.3em}
\end{algorithm}

%% file: table/blocks.tex
{\renewcommand{\arraystretch}{1.1}
\begin{table}[t]
\centering
\resizebox{.42\textwidth}{!}{
\begin{tabular}{cccc} 
\toprule
\#Partitions     & WRN-28 (12) & WRN-40 (18) & ResNet-50 (16)  \\ 
\toprule
\textbf{K=4} & 2,2,4,4     & 3,3,6,6     & 3,3,5,5         \\
\textbf{K=5} & 2,2,2,2,4   & 3,3,3,3,6   & 2,2,4,4,4       \\
\toprule
\end{tabular}}
\vspace{-.5em}
\caption{\textbf{Details of \# of blocks of each partition.} The list of numbers denotes the number of residual blocks for each part of the original networks, from the shallow to the deep parts sequentially. The values in parentheses are the total number of residual blocks.
}
\label{tab:blocks}
\vspace{-.2em}
\end{table}}

%% file: table/kernel_size.tex
{\renewcommand{\arraystretch}{1.1}
\begin{table}[t]
\centering
\resizebox{.48\textwidth}{!}{
\begin{tabular}{c|ccc|ccc} 
\toprule
(K=4) & \multicolumn{3}{c|}{\textbf{Kernel size= 1, padding=0}}       & \multicolumn{3}{c}{\textbf{Kernel size=3, padding=1}}         \\
\textbf{Arch} & \textbf{Avg. err} & \textbf{params~{\scriptsize $\uparrow$}}  & \textbf{Mem.} & \textbf{Avg. err} & \textbf{params~{\scriptsize $\uparrow$}} & \textbf{Mem.}  \\ 
\toprule
WRN-28    & 17.2          & 0.8\%       & 396       & {\cellcolor[rgb]{0.9,0.9,0.9}}\textbf{16.9}          & {\cellcolor[rgb]{0.9,0.9,0.9}}9.5\%       & {\cellcolor[rgb]{0.9,0.9,0.9}}404        \\
WRN-40    & 12.4         & 0.6\%       & 80        & {\cellcolor[rgb]{0.9,0.9,0.9}}\textbf{12.2}          & {\cellcolor[rgb]{0.9,0.9,0.9}}6.4\%       & {\cellcolor[rgb]{0.9,0.9,0.9}}80         \\
ResNet-50  & {\cellcolor[rgb]{0.9,0.9,0.9}}14.4          & {\cellcolor[rgb]{0.9,0.9,0.9}}11.8\%      & {\cellcolor[rgb]{0.9,0.9,0.9}}296       & \textbf{14.2}          & 142.2\%     & 394        \\
\toprule
\end{tabular}}
\vspace{-.5em}
\caption{\textbf{Kernel size in the conv layer.} We report the average error rate (\%), the increase rate of the model parameters compared to the original model (\%), and the total memory consumption (MB) including the model and activation sizes, based on the kernel size of the conv layer in meta networks.}
\label{tab:kernel}
\vspace{-.2em}
\end{table}}

%% file: table/transformation.tex
{\renewcommand{\arraystretch}{1.1}
\begin{table}[t]
\centering
\resizebox{.48\textwidth}{!}{
\begin{tabular}{cc|c|cccc} 
\toprule       
(K=4) &        &         & \multicolumn{4}{c}{\textbf{Transformations}}                                                                 \\
 \textbf{Dataset}   & \textbf{Arch}       & \textbf{EATA~\cite{eata}}     & \textbf{None}     & \textbf{+Color}    & \textbf{+Blur}    & \textbf{+Gray}    \\ 
\drule
CIFAR10-C  & WRN-40 & 13.0 & 12.5 & 12.3 & 12.3 & {\cellcolor[rgb]{0.9,0.9,0.9}}\textbf{12.2}  \\
CIFAR10-C  & WRN-28 & 18.6 & 17.8 & 17.4 & 17.2 & {\cellcolor[rgb]{0.9,0.9,0.9}}\textbf{16.9}  \\
CIFAR100-C & WRN-40 & 37.1 & 36.9 & 36.7 & 36.6 & {\cellcolor[rgb]{0.9,0.9,0.9}}\textbf{36.4}  \\
\toprule
\end{tabular}}
\vspace{-.5em}
\caption{\textbf{Ablation of the combination of transformations.} To warm up the meta networks, we use the following transformations in Pytorch: ColorJitter (Color), GaussianBlur (Blur), and RandomGrayscale (Gray). We report the average error rate (\%).
}
\label{tab:transformation}
\vspace{-.6em}
\end{table}}

%% file: table/arch_variants.tex
\begin{table*}[t!]
\centering
\subfloat[
\textbf{Visualization of meta networks}
\label{tab:vi_metanet}
]{
\centering
\begin{minipage}{0.35\linewidth}{\begin{center}

\includegraphics[width=0.92\linewidth]{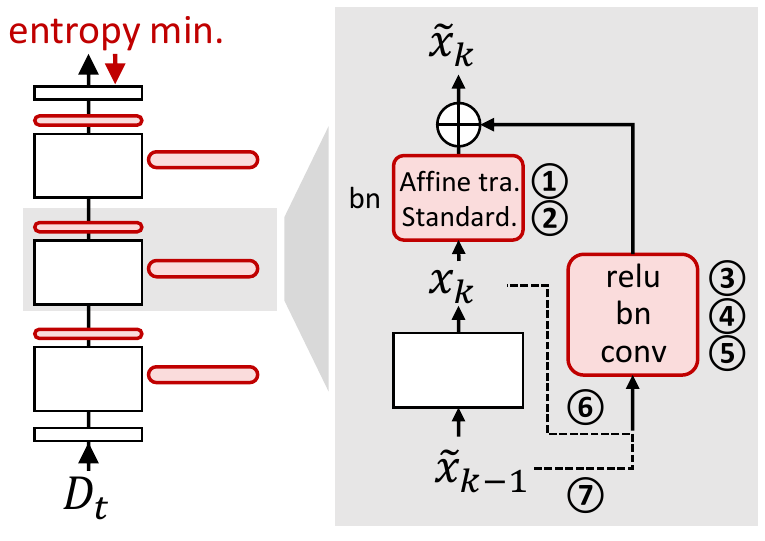}
\vspace{+.1em}
\end{center}

}\end{minipage}
}
\subfloat[
\textbf{Comparison of average error rate (\%) on continual TTA setup (K=5)}
\label{tab:meta_ablation}
]{
\begin{minipage}{0.64\linewidth}{\begin{center}

{\renewcommand{\arraystretch}{1.1}
\resizebox{.99\textwidth}{!}{
\huge
\begin{tabular}{c|ccccccc|ccc} 
\toprule
    (K=5)    &   &   &   &   &   &     &    & \multirow{2}{*}{\begin{tabular}[c]{@{}c@{}}~CIFAR10-C~ \\\textbf{WRN-28}\end{tabular}} & \multirow{2}{*}{\begin{tabular}[c]{@{}c@{}}~CIFAR10-C~ \\\textbf{WRN-40}\end{tabular}} & \multirow{2}{*}{\begin{tabular}[c]{@{}c@{}}~CIFAR100-C~ \\\textbf{WRN-40}\end{tabular}}  \\
\textbf{~~Variants~~} & \textbf{~~~\circled{1}~~~} & \textbf{~~~\circled{2}~~~} & \textbf{~~~\circled{3}~~~} & \textbf{~~~\circled{4}~~~} & \textbf{~~~\circled{5}~~~} & \textbf{~~~\circled{6}~~~} & \textbf{~~~\circled{7}~~~}      &    & &     \\ 
\drule
I       &  & $\checkmark$  & $\checkmark$  & $\checkmark$  & $\checkmark$  &  & $\checkmark$ & 19.9                                                                     & 15.4                                                                     & 39.2                                                                       \\
II      & $\checkmark$  &  & $\checkmark$  & $\checkmark$  &  $\checkmark$ & & $\checkmark$  & 18.6                                                                     & 13.4                                                                     & 38.0                                                                       \\
III       &  &  & $\checkmark$  & $\checkmark$  &  $\checkmark$ & & $\checkmark$  & 18.7                                                                     & 13.7                                                                     & 38.2                                                                       \\
IV        & $\checkmark$  &  $\checkmark$ &  &  $\checkmark$ &  $\checkmark$ & & $\checkmark$  & {18.6}        & {12.4}      & {36.7}                  \\
V       & $\checkmark$  & $\checkmark$  &  $\checkmark$ &  &  $\checkmark$ & & $\checkmark$  & 19.8                                                                     & 12.9                                                                     & 37.2                                                                       \\
VI      &  $\checkmark$ &  $\checkmark$ &  &  &  $\checkmark$ &  & $\checkmark$ & 32.3                                                                     & 14.5                                                                     & 51.8                                                                       \\
VII     & $\checkmark$  &  $\checkmark$ &  &  &  &  &  $\checkmark$ & 20.7                                                                     & 14.9                                                                     & 40.1                                                                       \\
XIII       & $\checkmark$  &  $\checkmark$ &  $\checkmark$ & $\checkmark$  & $\checkmark$ &  $\checkmark$  &  & {18.1}                                                                     & {12.6}                                                                     & {37.2}                                                                       \\
IX        &  &  & $\checkmark$  & $\checkmark$  & $\checkmark$  &  $\checkmark$  & & 60.6                                                                     & 73.3                                                                     & 77.2                                                                       \\
\rowcolor[rgb]{0.9,0.9,0.9} \textbf{Ours}        & $\checkmark$ & $\checkmark$  & $\checkmark$  &  $\checkmark$ & $\checkmark$  &     &  $\checkmark$  & \textbf{16.8}                                                                     & \textbf{12.1}                                                                     & \textbf{36.3}                                                                       \\
\toprule
\end{tabular}}}
\vspace{-.1em}

\end{center}}

\end{minipage}
}

\vspace{-.3em}
\caption{\textbf{Components of meta networks.} We conduct an ablation study on components of meta networks (\ie, {\footnotesize \textcircled{1} $\sim$ \textcircled{7}}). Here, {\footnotesize \textcircled{1}} and {\footnotesize \textcircled{2}} refer to affine transformation and standardization in a BN layer after the original networks. {\footnotesize \textcircled{3}$\sim$\textcircled{5}} and {\footnotesize \textcircled{6}$\sim$\textcircled{7}}, respectively, indicate modules in a convolution block and two kinds of inputs of it. In table (b), $\checkmark$ means applying the component to meta networks. 
}
\label{tab:add_arch_ablations}
\vspace{-1.em}
\end{table*}

%% file: table/main_task_loss.tex
{\renewcommand{\arraystretch}{1.05}
\begin{table}[t]
\centering
{\small
\resizebox{.47\textwidth}{!}{
\begin{tabular}{cc|c|ccc} 
\toprule
(K=5) & & & \multicolumn{3}{c}{\textbf{Ours}}  \\
\textbf{Dataset}                     & \textbf{Arch}     & \textbf{EATA}~\cite{eata} & ~~{\small $\loss^{1}$}~~ & ~~{\small $\loss^{2}$}~~ & {\cellcolor[rgb]{0.9,0.9,0.9}} ~~{\small $\loss^{3}$}~~ \\ 
\drule
\multirow{3}{*}{CIFAR10-C}  & WRN-28    & 18.6  & 17.3   & \underbar{16.9}   & {\cellcolor[rgb]{0.9,0.9,0.9}}\textbf{16.9}    \\
                            & WRN-40    & 13.0  & \underbar{12.2}   & 12.3   & {\cellcolor[rgb]{0.9,0.9,0.9}}\textbf{12.1}    \\
                            & Resnet-50 & 14.2  & 15.0   & \underbar{14.3}   & {\cellcolor[rgb]{0.9,0.9,0.9}}\textbf{14.1}    \\
\hline
\multirow{2}{*}{CIFAR100-C} & WRN-40    & 37.1  & 36.5   & \underbar{36.4}   & {\cellcolor[rgb]{0.9,0.9,0.9}}\textbf{36.3}    \\
                            & Resnet-50 & 39.9  & 40.7   & \textbf{38.8}   & {\cellcolor[rgb]{0.9,0.9,0.9}}\underbar{39.4}    \\
\toprule
\end{tabular}}}
\vspace{-.5em}
\caption{\textbf{Ablation study of main task loss.} We compare the average error rate (\%) of three types of adaptation losses.}
\label{tab:mainloss}
\vspace{-.5em}
\end{table}}

%% file: table/reg_loss.tex
{\renewcommand{\arraystretch}{1.1}
\begin{table}[t]
\centering
\resizebox{.43\textwidth}{!}{
\begin{tabular}{cc|cc} 
\toprule
(K=5)                       &               & \multicolumn{2}{c}{\textbf{Ours}}                       \\
\textbf{Dataset}            & \textbf{Arch} & MSE loss (\equref{eq:mse}) & {\cellcolor[rgb]{0.9,0.9,0.9}}L1 loss (Eq.~(\textcolor{red}{4})) \\
\drule
\multirow{3}{*}{CIFAR10-C}  & WRN-28        &  16.9    & {\cellcolor[rgb]{0.9,0.9,0.9}}16.9         \\
                            & WRN-40        &  12.3    & {\cellcolor[rgb]{0.9,0.9,0.9}}\textbf{12.1}         \\
                            & Resnet-50     &  14.1    & {\cellcolor[rgb]{0.9,0.9,0.9}}14.1         \\ 
\hline
\multirow{2}{*}{CIFAR100-C} & WRN-40        &   36.6   & {\cellcolor[rgb]{0.9,0.9,0.9}}\textbf{36.3}         \\
                            & Resnet-50     &   39.5    & {\cellcolor[rgb]{0.9,0.9,0.9}}\textbf{39.4}         \\
\toprule
\end{tabular}}
\vspace{-.5em}
\caption{\textbf{Ablation study of loss function of our regularization.} We present the average error (\%) according to two types of loss functions for self-distilled regularization.}
\label{tab:regloss}
\vspace{-1.0em}
\end{table}}

%% file: table/reg_scale.tex
\begin{table}[!ht]
\vspace{-0.3em}
\centering
\resizebox{.47\textwidth}{!}{
\begin{tabular}{c|ccccc|cc}
\textbf{Round $\setminus \lambda$}   & 0     & 0.1   & 0.5   & 1     & 2     & 5     & 10     \\ 
\hline
1  & 36.31 & 36.30  & \textbf{36.29} & 36.56 & 37.20 & 38.41 & 39.58  \\
10 & 55.47 & 43.83 & 36.42 & 36.14 & 36.48 & 37.47 & 38.95 
\end{tabular}}
\vspace{-0.3em}
\end{table}

%% file: table/cifar10_all.tex
{\renewcommand{\arraystretch}{1.}
\begin{table*}[t!]
\centering
\resizebox{.99\textwidth}{!}{
\begin{tabular}{c|l|ccccccccccccccc|cc} 
\toprule
Time       &                                                & \multicolumn{15}{l|}{$t\xrightarrow{\hspace*{17.2cm}}$}   &                                          &                                          \\
\textbf{Arch} & \textbf{Method}                                & \textbf{Gaus.}                           & \textbf{Shot}                            & \textbf{Impu.}                           & \textbf{Defo.}                           & \textbf{Glas.}                           & \textbf{Moti.}                           & \textbf{Zoom}                            & \textbf{Snow}                            & \textbf{Fros.}                           & \textbf{Fog}                             & \textbf{Brig.}                          & \textbf{Cont.}                           & \textbf{Elas.}                           & \textbf{Pixe.}                           & \textbf{Jpeg}                            & \textbf{Avg. err}                        & \textbf{Mem.}                            \\ 
\drule
\multirow{11}{*}{\begin{tabular}[c]{@{}c@{}}WRN-40\\(AugMix)\end{tabular}}        & Source                                         & 44.3                                     & 37.0                                     & 44.8                                     & 30.6                                     & 43.9                                     & 32.6                                     & 29.4                                     & 23.9                                     & 30.1                                     & 39.7                                     & 12.9                                    & 66.4                                     & 32.7                                     & 58.4                                     & 23.5                                     & 36.7                                     & 11                                       \\
      & tBN~\cite{tbn}                                            & 19.5                                     & 17.6                                     & 23.8                                     & 9.6                                      & 23.1                                     & 11.1                                     & 10.3                                     & 13.4                                     & 14.2                                     & 15.0                                     & 8.0                                     & 13.9                                     & 17.3                                     & 16.0                                     & 18.8                                     & 15.4                                     & 11                                       \\
              & Single do. TENT~\cite{tent}                                    & 16.4                                     & 13.9                                     & 19.1                                     & 8.3                                      & 19.1                                     & 9.3                                      & 8.6                                      & 10.9                                     & 11.3                                     & 12.0                                     & 6.9                                     & 11.6                                     & 14.6                                     & 12.2                                     & 15.6                                     & 12.7                                     & 188                                      \\
              & TENT continual~\cite{tent}                                 & 16.4                                     & 12.2                                     & 17.1                                     & 9.1                                      & 18.7                                     & 11.4                                     & 10.4                                     & 12.7                                     & 12.4                                     & 14.8                                     & 10.1                                    & 13.0                                     & 17.0                                     & 13.3                                     & 19.0                                     & 13.3                                     & 188                                      \\
              & TTT++~\cite{ttt++}                                         & 19.1                                     & 16.9                                     & 22.2                                     & 9.3                                      & 21.6                                     & 10.8                                     & 9.8                                      & 12.7                                     & 13.1                                     & 14.3                                     & 7.8                                     & 13.9                                     & 15.9                                     & 14.2                                     & 17.2                                     & 14.6                                     & 391                                      \\
              & SWRNSP~\cite{swr}                                         & 15.9                                     & 13.3                                     & 18.2                                     & 8.4                                      & 18.5                                     & 9.5                                      & 8.6                                      & 11.0                                     & 10.2                                     & 11.7                                     & 7.0                                     & 8.1                                      & 14.6                                     & 11.3                                     & 15.1                                     & 12.1                                     & 400                                      \\
              & NOTE~\cite{note}                                           & 19.6                                     & 16.4                                     & 19.9                                     & 9.4                                      & 20.3                                     & 10.3                                     & 10.1                                     & 11.6                                     & 10.6                                     & 13.3                                     & 7.9                                     & 7.7                                      & 15.4                                     & 12.0                                     & 17.3                                     & 13.4                                     & 188                                      \\
              & EATA~\cite{eata}                                           & 15.2                                     & 13.1                                     & 17.5                                     & 9.5                                      & 19.9                                     & 11.6                                     & 9.3                                      & 11.4                                     & 11.5                                     & 12.4                                     & 7.8                                     & 11.1                                     & 16.1                                     & 12.2                                     & 16.1                                     & 13.0                                     & 188                                      \\
              & CoTTA~\cite{cotta}                                          & 15.6                                     & 13.6                                     & 17.3                                     & 9.8                                      & 19.0                                     & 11.0                                     & 10.2                                     & 13.5                                     & 12.6                                     & 17.4                                     & 7.8                                     & 17.3                                     & 16.2                                     & 12.9                                     & 16.0                                     & 14.0                                     & 409                                      \\
              & {\cellcolor[rgb]{0.9,0.9,0.9}}Ours (K=4) & {\cellcolor[rgb]{0.9,0.9,0.9}}16.1 & {\cellcolor[rgb]{0.9,0.9,0.9}}13.2 & {\cellcolor[rgb]{0.9,0.9,0.9}}18.3 & {\cellcolor[rgb]{0.9,0.9,0.9}}8.0  & {\cellcolor[rgb]{0.9,0.9,0.9}}18.3 & {\cellcolor[rgb]{0.9,0.9,0.9}}9.3  & {\cellcolor[rgb]{0.9,0.9,0.9}}8.6  & {\cellcolor[rgb]{0.9,0.9,0.9}}10.5 & {\cellcolor[rgb]{0.9,0.9,0.9}}10.1 & {\cellcolor[rgb]{0.9,0.9,0.9}}12.2 & {\cellcolor[rgb]{0.9,0.9,0.9}}6.8 & {\cellcolor[rgb]{0.9,0.9,0.9}}11.3 & {\cellcolor[rgb]{0.9,0.9,0.9}}14.5 & {\cellcolor[rgb]{0.9,0.9,0.9}}11.0 & {\cellcolor[rgb]{0.9,0.9,0.9}}14.8 & {\cellcolor[rgb]{0.9,0.9,0.9}}12.2 & {\cellcolor[rgb]{0.9,0.9,0.9}}80   \\
              & {\cellcolor[rgb]{0.9,0.9,0.9}}Ours (K=5) & {\cellcolor[rgb]{0.9,0.9,0.9}}15.9 & {\cellcolor[rgb]{0.9,0.9,0.9}}12.6 & {\cellcolor[rgb]{0.9,0.9,0.9}}17.2 & {\cellcolor[rgb]{0.9,0.9,0.9}}8.2  & {\cellcolor[rgb]{0.9,0.9,0.9}}18.4 & {\cellcolor[rgb]{0.9,0.9,0.9}}9.3  & {\cellcolor[rgb]{0.9,0.9,0.9}}8.6  & {\cellcolor[rgb]{0.9,0.9,0.9}}10.6 & {\cellcolor[rgb]{0.9,0.9,0.9}}10.4 & {\cellcolor[rgb]{0.9,0.9,0.9}}12.4 & {\cellcolor[rgb]{0.9,0.9,0.9}}6.7 & {\cellcolor[rgb]{0.9,0.9,0.9}}11.7 & {\cellcolor[rgb]{0.9,0.9,0.9}}14.3 & {\cellcolor[rgb]{0.9,0.9,0.9}}11.3 & {\cellcolor[rgb]{0.9,0.9,0.9}}14.9 & {\cellcolor[rgb]{0.9,0.9,0.9}}12.1 & {\cellcolor[rgb]{0.9,0.9,0.9}}92   \\
\hline
\multirow{11}{*}{WRN-28}        & Source                                         & 72.3                                     & 65.7                                     & 72.9                                     & 46.9                                     & 54.3                                     & 34.8                                     & 42.0                                     & 25.1                                     & 41.3                                     & 26.0                                     & 9.3                                     & 46.7                                     & 26.6                                     & 58.5                                     & 30.3                                     & 43.5                                     & 58                                       \\
              & tBN~\cite{tbn}                                            & 28.6                                     & 26.8                                     & 37.0                                     & 13.2                                     & 35.4                                     & 14.4                                     & 12.6                                     & 18.0                                     & 18.2                                     & 16.0                                     & 8.6                                     & 13.3                                     & 24.0                                     & 20.3                                     & 27.8                                     & 20.9                                     & 58                                       \\
              & Single do. TENT~\cite{tent}                                    & 25.2                                     & 23.8                                     & 33.5                                     & 12.8                                     & 32.3                                     & 14.1                                     & 11.7                                     & 16.4                                     & 17.0                                     & 14.4                                     & 8.4                                     & 12.2                                     & 22.8                                     & 18.0                                     & 24.8                                     & 19.2                                     & 646                                      \\
              & Continual TENT~\cite{tent}                                 & 25.2                                     & 20.8                                     & 29.8                                     & 14.4                                     & 31.5                                     & 15.4                                     & 14.2                                     & 18.8                                     & 17.5                                     & 17.3                                     & 10.9                                    & 14.9                                     & 23.6                                     & 20.2                                     & 25.6                                     & 20.0                                     & 646                                      \\
              & TTT++~\cite{ttt++}                                         & 27.9                                     & 25.8                                     & 35.8                                     & 13.0                                     & 34.3                                     & 14.2                                     & 12.2                                     & 17.4                                     & 17.6                                     & 15.5                                     & 8.6                                     & 13.1                                     & 23.1                                     & 19.6                                     & 26.6                                     & 20.3                                     & 1405                                     \\
              & SWRNSP~\cite{swr}                                         & 24.6                                     & 20.5                                     & 29.3                                     & 12.4                                     & 31.1                                     & 13.0                                     & 11.3                                     & 15.3                                     & 14.7                                     & 11.7                                     & 7.8                                     & 9.3                                      & 21.5                                     & 15.6                                     & 20.3                                     & 17.2                                     & 1551                                     \\
              & NOTE~\cite{note}                                           & 30.4                                     & 26.7                                     & 34.6                                     & 13.6                                     & 36.3                                     & 13.7                                     & 13.9                                     & 17.2                                     & 15.8                                     & 15.2                                     & 9.1                                     & 7.5                                      & 24.1                                     & 18.4                                     & 25.9                                     & 20.2                                     & 646                                      \\
              & EATA~\cite{eata}                                           & 23.8                                     & 18.8                                     & 27.3                                     & 13.9                                     & 29.7                                     & 16.0                                     & 13.3                                     & 18.0                                     & 16.9                                     & 15.7                                     & 10.5                                    & 12.2                                     & 22.9                                     & 17.1                                     & 23.0                                     & 18.6                                     & 646                                      \\
              & CoTTA~\cite{cotta}                                          & 24.6                                     & 21.6                                     & 26.5                                     & 12.1                                     & 28.0                                     & 13.0                                     & 10.9                                     & 15.3                                     & 14.6                                     & 13.6                                     & 8.1                                     & 12.2                                     & 20.0                                     & 14.9                                     & 19.5                                     & 17.0                                     & 1697                                     \\
              & {\cellcolor[rgb]{0.9,0.9,0.9}}Ours (K=4) & {\cellcolor[rgb]{0.9,0.9,0.9}}23.5 & {\cellcolor[rgb]{0.9,0.9,0.9}}19.0 & {\cellcolor[rgb]{0.9,0.9,0.9}}26.6 & {\cellcolor[rgb]{0.9,0.9,0.9}}11.5 & {\cellcolor[rgb]{0.9,0.9,0.9}}30.6 & {\cellcolor[rgb]{0.9,0.9,0.9}}13.1 & {\cellcolor[rgb]{0.9,0.9,0.9}}10.9 & {\cellcolor[rgb]{0.9,0.9,0.9}}15.2 & {\cellcolor[rgb]{0.9,0.9,0.9}}14.5 & {\cellcolor[rgb]{0.9,0.9,0.9}}13.1 & {\cellcolor[rgb]{0.9,0.9,0.9}}7.8 & {\cellcolor[rgb]{0.9,0.9,0.9}}11.4 & {\cellcolor[rgb]{0.9,0.9,0.9}}20.9 & {\cellcolor[rgb]{0.9,0.9,0.9}}15.4 & {\cellcolor[rgb]{0.9,0.9,0.9}}20.8 & {\cellcolor[rgb]{0.9,0.9,0.9}}16.9 & {\cellcolor[rgb]{0.9,0.9,0.9}}404  \\
              & {\cellcolor[rgb]{0.9,0.9,0.9}}Ours (K=5) & {\cellcolor[rgb]{0.9,0.9,0.9}}23.8 & {\cellcolor[rgb]{0.9,0.9,0.9}}18.7 & {\cellcolor[rgb]{0.9,0.9,0.9}}25.7 & {\cellcolor[rgb]{0.9,0.9,0.9}}11.5 & {\cellcolor[rgb]{0.9,0.9,0.9}}29.8 & {\cellcolor[rgb]{0.9,0.9,0.9}}13.3 & {\cellcolor[rgb]{0.9,0.9,0.9}}11.3 & {\cellcolor[rgb]{0.9,0.9,0.9}}15.3 & {\cellcolor[rgb]{0.9,0.9,0.9}}15.0 & {\cellcolor[rgb]{0.9,0.9,0.9}}13.0 & {\cellcolor[rgb]{0.9,0.9,0.9}}7.9 & {\cellcolor[rgb]{0.9,0.9,0.9}}11.3 & {\cellcolor[rgb]{0.9,0.9,0.9}}20.2 & {\cellcolor[rgb]{0.9,0.9,0.9}}15.1 & {\cellcolor[rgb]{0.9,0.9,0.9}}20.5 & {\cellcolor[rgb]{0.9,0.9,0.9}}16.8 & {\cellcolor[rgb]{0.9,0.9,0.9}}471  \\
\hline
\multirow{10}{*}{Resnet-50}     & Source                                         & 65.6                                     & 60.7                                     & 74.4                                     & 28.9                                     & 79.9                                     & 46.0                                     & 25.7                                     & 35.0                                     & 49.4                                     & 54.7                                     & 13.0                                    & 83.2                                     & 41.2                                     & 46.7                                     & 27.7                                     & 48.8                                     & 91                                       \\
              & tBN~\cite{tbn}                                            & 18.0                                     & 17.2                                     & 29.3                                     & 10.7                                     & 27.2                                     & 15.5                                     & 8.9                                      & 16.7                                     & 14.6                                     & 21.0                                     & 9.3                                     & 12.7                                     & 20.9                                     & 12.4                                     & 14.8                                     & 16.6                                     & 91                                       \\
              & Single do. TENT~\cite{tent}                                    & 16.6                                     & 15.7                                     & 25.7                                     & 10.0                                     & 24.8                                     & 13.8                                     & 8.3                                      & 14.9                                     & 13.8                                     & 17.6                                     & 8.7                                     & 10.0                                     & 19.1                                     & 11.5                                     & 13.8                                     & 15.0                                     & 925                                      \\
              & TENT continual~\cite{tent}                                 & 16.6                                     & 14.4                                     & 22.9                                     & 10.4                                     & 22.6                                     & 13.4                                     & 10.3                                     & 15.8                                     & 14.6                                     & 18.0                                     & 10.5                                    & 11.7                                     & 18.4                                     & 13.1                                     & 15.3                                     & 15.2                                     & 925                                      \\
              & TTT++~\cite{ttt++}                                          & 18.2                                     & 16.9                                     & 28.7                                     & 10.5                                     & 26.5                                     & 14.5                                     & 8.9                                      & 16.5                                     & 14.5                                     & 20.9                                     & 9.0                                     & 9.0                                      & 20.4                                     & 12.3                                     & 14.7                                     & 16.1                                     & 1877                                     \\
              & SWRNSP~\cite{swr}                                         & 17.3                                     & 16.1                                     & 26.1                                     & 10.6                                     & 25.6                                     & 14.1                                     & 8.7                                      & 15.6                                     & 13.6                                     & 18.6                                     & 8.8                                     & 10.0                                     & 19.3                                     & 12.0                                     & 14.2                                     & 15.4                                     & 1971                                     \\
              & EATA~\cite{eata}                                           & 17.2                                     & 14.9                                     & 23.6                                     & 10.2                                     & 23.3                                     & 13.2                                     & 8.5                                      & 14.0                                     & 12.5                                     & 16.6                                     & 8.6                                     & 9.4                                      & 17.2                                     & 11.0                                     & 12.7                                     & 14.2                                     & 925                                      \\
              & CoTTA~\cite{cotta}                                          & 16.2                                     & 15.0                                     & 21.2                                     & 10.4                                     & 22.8                                     & 13.9                                     & 8.4                                      & 15.1                                     & 12.9                                     & 19.8                                     & 8.6                                     & 11.3                                     & 17.5                                     & 10.5                                     & 12.2                                     & 14.4                                     & 2066                                     \\
              & {\cellcolor[rgb]{0.9,0.9,0.9}}Ours (K=4) & {\cellcolor[rgb]{0.9,0.9,0.9}}16.5 & {\cellcolor[rgb]{0.9,0.9,0.9}}14.5 & {\cellcolor[rgb]{0.9,0.9,0.9}}24.3 & {\cellcolor[rgb]{0.9,0.9,0.9}}9.7  & {\cellcolor[rgb]{0.9,0.9,0.9}}23.7 & {\cellcolor[rgb]{0.9,0.9,0.9}}13.3 & {\cellcolor[rgb]{0.9,0.9,0.9}}8.8  & {\cellcolor[rgb]{0.9,0.9,0.9}}14.7 & {\cellcolor[rgb]{0.9,0.9,0.9}}12.9 & {\cellcolor[rgb]{0.9,0.9,0.9}}17.0 & {\cellcolor[rgb]{0.9,0.9,0.9}}9.1 & {\cellcolor[rgb]{0.9,0.9,0.9}}9.4  & {\cellcolor[rgb]{0.9,0.9,0.9}}17.6 & {\cellcolor[rgb]{0.9,0.9,0.9}}11.4 & {\cellcolor[rgb]{0.9,0.9,0.9}}13.1 & {\cellcolor[rgb]{0.9,0.9,0.9}}14.4 & {\cellcolor[rgb]{0.9,0.9,0.9}}296  \\
              & {\cellcolor[rgb]{0.9,0.9,0.9}}Ours (K=5) & {\cellcolor[rgb]{0.9,0.9,0.9}}16.6 & {\cellcolor[rgb]{0.9,0.9,0.9}}14.4 & {\cellcolor[rgb]{0.9,0.9,0.9}}23.6 & {\cellcolor[rgb]{0.9,0.9,0.9}}9.8  & {\cellcolor[rgb]{0.9,0.9,0.9}}23.4 & {\cellcolor[rgb]{0.9,0.9,0.9}}12.7 & {\cellcolor[rgb]{0.9,0.9,0.9}}8.6  & {\cellcolor[rgb]{0.9,0.9,0.9}}14.5 & {\cellcolor[rgb]{0.9,0.9,0.9}}12.6 & {\cellcolor[rgb]{0.9,0.9,0.9}}16.6 & {\cellcolor[rgb]{0.9,0.9,0.9}}8.7 & {\cellcolor[rgb]{0.9,0.9,0.9}}9.0  & {\cellcolor[rgb]{0.9,0.9,0.9}}17.0 & {\cellcolor[rgb]{0.9,0.9,0.9}}11.3 & {\cellcolor[rgb]{0.9,0.9,0.9}}12.6 & {\cellcolor[rgb]{0.9,0.9,0.9}}14.1 & {\cellcolor[rgb]{0.9,0.9,0.9}}498  \\
\toprule
\end{tabular}}
\vspace{-.5em}
\caption{\textbf{Comparison of error rate ($\%$) on CIFARC10-C with severity level 5.} We conduct experiments on continual TTA setup. Avg. err means the average error rate (\%) of all 15 corruptions, and Mem. denotes total memory consumption, including model parameter sizes and activations. WRN refers to WideResNet. The implementation details of the baselines are described in \secref{sec:ttaworks}.}
\label{tab:cifar10_all} 
\vspace{-0.0em}
\end{table*}}

%% file: table/cifar100_all.tex
{\renewcommand{\arraystretch}{1.}
\begin{table*}[t!]
\centering
\resizebox{.99\textwidth}{!}{
\begin{tabular}{c|l|ccccccccccccccc|cc} 
\toprule
Time       &                                                & \multicolumn{15}{l|}{$t\xrightarrow{\hspace*{17.2cm}}$}   &                                          &                                          \\
\textbf{Arch} & \textbf{Method}                                & \textbf{Gaus.}                           & \textbf{Shot}                            & \textbf{Impu.}                           & \textbf{Defo.}                           & \textbf{Glas.}                           & \textbf{Moti.}                           & \textbf{Zoom}                            & \textbf{Snow}                            & \textbf{Fros.}                           & \textbf{Fog}                             & \textbf{Brig.}                          & \textbf{Cont.}                           & \textbf{Elas.}                           & \textbf{Pixe.}                           & \textbf{Jpeg}                            & \textbf{Avg. err}                        & \textbf{Mem.}                            \\ 
\drule
\multirow{11}{*}{\begin{tabular}[c]{@{}l@{}}WRN-40\\(AugMix)\end{tabular}}    & Source                                         & 80.1                                     & 77.0                                     & 76.4                                     & 59.9                                     & 77.6                                     & 64.2                                     & 59.3                                     & 64.8                                     & 71.3                                     & 78.3                                     & 48.1                                     & 83.4                                     & 65.8                                     & 80.4                                     & 59.2                                     & 69.7                                     & 11                                       \\ 
  & tBN~\cite{tbn}                                            & 45.9                                     & 45.6                                     & 48.2                                     & 33.6                                     & 47.9                                     & 34.5                                     & 34.1                                     & 40.3                                     & 40.4                                     & 47.1                                     & 31.7                                     & 39.7                                     & 42.7                                     & 39.2                                     & 45.6                                     & 41.1                                     & 11                                       \\
          & Single do. TENT~\cite{tent}                                    & 41.2                                     & 40.6                                     & 42.2                                     & 30.9                                     & 43.4                                     & 31.8                                     & 30.6                                     & 35.3                                     & 36.2                                     & 40.1                                     & 28.5                                     & 35.5                                     & 39.1                                     & 33.9                                     & 41.7                                     & 36.7                                     & 188                                      \\
          & continual TENT~\cite{tent}                                 & 41.2                                     & 38.2                                     & 41.0                                     & 32.9                                     & 43.9                                     & 34.9                                     & 33.2                                     & 37.7                                     & 37.2                                     & 41.5                                     & 33.2                                     & 37.2                                     & 41.1                                     & 35.9                                     & 45.1                                     & 38.3                                     & 188                                      \\
          & TTT++~\cite{ttt++}                                         & 46.0                                     & 45.4                                     & 48.2                                     & 33.5                                     & 47.7                                     & 34.4                                     & 33.8                                     & 39.9                                     & 40.2                                     & 47.1                                     & 31.8                                     & 39.7                                     & 42.5                                     & 38.9                                     & 45.5                                     & 41.0                                     & 391                                      \\
          & SWRNSP~\cite{swr}                                         & 42.4                                     & 40.9                                     & 42.7                                     & 30.6                                     & 43.9                                     & 31.7                                     & 31.3                                     & 36.1                                     & 36.2                                     & 41.5                                     & 28.7                                     & 34.1                                     & 39.2                                     & 33.6                                     & 41.3                                     & 36.6                                     & 400                                      \\
          & NOTE~\cite{note}                                           & 50.9                                     & 47.4                                     & 49.0                                     & 37.3                                     & 49.6                                     & 37.3                                     & 37.0                                     & 41.3                                     & 39.9                                     & 47.0                                     & 35.2                                     & 34.7                                     & 45.2                                     & 40.9                                     & 49.9                                     & 42.8                                     & 188                                      \\
          & EATA~\cite{eata}                                           & 41.6                                     & 39.9                                     & 41.2                                     & 31.7                                     & 44.0                                     & 32.4                                     & 31.9                                     & 36.2                                     & 36.8                                     & 39.7                                     & 29.1                                     & 34.4                                     & 39.9                                     & 34.2                                     & 42.2                                     & 37.1                                     & 188                                      \\
          & CoTTA~\cite{cotta}                                          & 43.5                                     & 41.7                                     & 43.7                                     & 32.2                                     & 43.7                                     & 32.8                                     & 32.2                                     & 38.5                                     & 37.6                                     & 45.9                                     & 29.0                                     & 38.1                                     & 39.2                                     & 33.8                                     & 39.4                                     & 38.1                                     & 409                                      \\
          & {\cellcolor[rgb]{0.9,0.9,0.9}}Ours (K=4) & {\cellcolor[rgb]{0.9,0.9,0.9}}42.7 & {\cellcolor[rgb]{0.9,0.9,0.9}}39.6 & {\cellcolor[rgb]{0.9,0.9,0.9}}42.4 & {\cellcolor[rgb]{0.9,0.9,0.9}}31.4 & {\cellcolor[rgb]{0.9,0.9,0.9}}42.9 & {\cellcolor[rgb]{0.9,0.9,0.9}}31.9 & {\cellcolor[rgb]{0.9,0.9,0.9}}30.8 & {\cellcolor[rgb]{0.9,0.9,0.9}}35.1 & {\cellcolor[rgb]{0.9,0.9,0.9}}34.8 & {\cellcolor[rgb]{0.9,0.9,0.9}}40.7 & {\cellcolor[rgb]{0.9,0.9,0.9}}28.1 & {\cellcolor[rgb]{0.9,0.9,0.9}}35.0 & {\cellcolor[rgb]{0.9,0.9,0.9}}37.5 & {\cellcolor[rgb]{0.9,0.9,0.9}}32.1 & {\cellcolor[rgb]{0.9,0.9,0.9}}40.5 & {\cellcolor[rgb]{0.9,0.9,0.9}}36.4 & {\cellcolor[rgb]{0.9,0.9,0.9}}80   \\
          & {\cellcolor[rgb]{0.9,0.9,0.9}}Ours (K=5) & {\cellcolor[rgb]{0.9,0.9,0.9}}41.8 & {\cellcolor[rgb]{0.9,0.9,0.9}}39.0 & {\cellcolor[rgb]{0.9,0.9,0.9}}41.9 & {\cellcolor[rgb]{0.9,0.9,0.9}}31.2 & {\cellcolor[rgb]{0.9,0.9,0.9}}42.7 & {\cellcolor[rgb]{0.9,0.9,0.9}}32.5 & {\cellcolor[rgb]{0.9,0.9,0.9}}31.0 & {\cellcolor[rgb]{0.9,0.9,0.9}}35.0 & {\cellcolor[rgb]{0.9,0.9,0.9}}35.0 & {\cellcolor[rgb]{0.9,0.9,0.9}}39.9 & {\cellcolor[rgb]{0.9,0.9,0.9}}28.8 & {\cellcolor[rgb]{0.9,0.9,0.9}}34.5 & {\cellcolor[rgb]{0.9,0.9,0.9}}37.5 & {\cellcolor[rgb]{0.9,0.9,0.9}}32.8 & {\cellcolor[rgb]{0.9,0.9,0.9}}40.5 & {\cellcolor[rgb]{0.9,0.9,0.9}}36.3 & {\cellcolor[rgb]{0.9,0.9,0.9}}92   \\ 
\hline
\multirow{10}{*}{Resnet-50} & Source                                         & 84.7                                     & 83.5                                     & 93.3                                     & 59.6                                     & 92.5                                     & 71.9                                     & 54.8                                     & 66.6                                     & 77.6                                     & 81.8                                     & 44.3                                     & 91.2                                     & 72.2                                     & 76.6                                     & 56.5                                     & 73.8                                     & 91                                       \\
          & tBN~\cite{tbn}                                            & 48.1                                     & 46.7                                     & 60.6                                     & 35.1                                     & 58.0                                     & 41.8                                     & 33.2                                     & 47.3                                     & 43.5                                     & 54.9                                     & 33.5                                     & 35.3                                     & 49.8                                     & 38.4                                     & 40.8                                     & 44.5                                     & 91                                       \\
          & Single do. TENT~\cite{tent}                                    & 44.1                                     & 42.7                                     & 53.9                                     & 32.6                                     & 52.0                                     & 37.5                                     & 30.5                                     & 43.4                                     & 40.2                                     & 45.7                                     & 30.4                                     & 31.4                                     & 45.1                                     & 35.0                                     & 37.6                                     & 40.1                                     & 926                                      \\
          & continual TENT~\cite{tent}                                 & 44.0                                     & 40.1                                     & 49.9                                     & 34.7                                     & 50.6                                     & 40.0                                     & 33.6                                     & 47.0                                     & 45.7                                     & 53.4                                     & 42.5                                     & 46.2                                     & 56.1                                     & 51.2                                     & 53.3                                     & 45.9                                     & 926                                      \\
          & TTT++~\cite{ttt++}                                          & 48.1                                     & 46.5                                     & 60.8                                     & 35.1                                     & 57.8                                     & 41.6                                     & 32.9                                     & 46.8                                     & 43.3                                     & 55.0                                     & 33.3                                     & 34.0                                     & 50.0                                     & 38.1                                     & 40.6                                     & 44.2                                     & 1876                                     \\
          & SWRNSP~\cite{swr}                                         & 48.3                                     & 46.5                                     & 60.5                                     & 35.1                                     & 57.9                                     & 41.7                                     & 32.9                                     & 47.1                                     & 43.5                                     & 54.7                                     & 33.5                                     & 35.1                                     & 49.9                                     & 38.3                                     & 40.7                                     & 44.1                                     & 1970                                     \\
          & EATA~\cite{eata}                                           & 44.8                                     & 41.9                                     & 52.6                                     & 33.0                                     & 51.1                                     & 37.8                                     & 30.3                                     & 43.0                                     & 40.1                                     & 45.1                                     & 30.1                                     & 31.8                                     & 45.2                                     & 35.2                                     & 37.4                                     & 39.9                                     & 926                                      \\
          & CoTTA~\cite{cotta}                                          & 43.6                                     & 42.8                                     & 50.4                                     & 34.2                                     & 51.6                                     & 39.2                                     & 31.4                                     & 43.4                                     & 39.6                                     & 47.4                                     & 31.3                                     & 32.2                                     & 43.4                                     & 35.8                                     & 36.7                                     & 40.2                                     & 2064                                     \\
          & {\cellcolor[rgb]{0.9,0.9,0.9}}Ours (K=4) & {\cellcolor[rgb]{0.9,0.9,0.9}}44.8 & {\cellcolor[rgb]{0.9,0.9,0.9}}40.3 & {\cellcolor[rgb]{0.9,0.9,0.9}}49.2 & {\cellcolor[rgb]{0.9,0.9,0.9}}32.3 & {\cellcolor[rgb]{0.9,0.9,0.9}}50.1 & {\cellcolor[rgb]{0.9,0.9,0.9}}36.3 & {\cellcolor[rgb]{0.9,0.9,0.9}}29.5 & {\cellcolor[rgb]{0.9,0.9,0.9}}41.0 & {\cellcolor[rgb]{0.9,0.9,0.9}}39.9 & {\cellcolor[rgb]{0.9,0.9,0.9}}44.6 & {\cellcolor[rgb]{0.9,0.9,0.9}}31.5 & {\cellcolor[rgb]{0.9,0.9,0.9}}33.7 & {\cellcolor[rgb]{0.9,0.9,0.9}}45.3 & {\cellcolor[rgb]{0.9,0.9,0.9}}36.3 & {\cellcolor[rgb]{0.9,0.9,0.9}}37.7 & {\cellcolor[rgb]{0.9,0.9,0.9}}39.5 & {\cellcolor[rgb]{0.9,0.9,0.9}}296  \\
          & {\cellcolor[rgb]{0.9,0.9,0.9}}Ours (K=5) & {\cellcolor[rgb]{0.9,0.9,0.9}}44.9 & {\cellcolor[rgb]{0.9,0.9,0.9}}40.4 & {\cellcolor[rgb]{0.9,0.9,0.9}}48.9 & {\cellcolor[rgb]{0.9,0.9,0.9}}32.7 & {\cellcolor[rgb]{0.9,0.9,0.9}}49.7 & {\cellcolor[rgb]{0.9,0.9,0.9}}36.9 & {\cellcolor[rgb]{0.9,0.9,0.9}}29.3 & {\cellcolor[rgb]{0.9,0.9,0.9}}40.8 & {\cellcolor[rgb]{0.9,0.9,0.9}}39.0 & {\cellcolor[rgb]{0.9,0.9,0.9}}44.4 & {\cellcolor[rgb]{0.9,0.9,0.9}}31.1 & {\cellcolor[rgb]{0.9,0.9,0.9}}33.6 & {\cellcolor[rgb]{0.9,0.9,0.9}}44.0 & {\cellcolor[rgb]{0.9,0.9,0.9}}35.7 & {\cellcolor[rgb]{0.9,0.9,0.9}}37.8 & {\cellcolor[rgb]{0.9,0.9,0.9}}39.3 & {\cellcolor[rgb]{0.9,0.9,0.9}}498  \\
\toprule
\end{tabular}}
\vspace{-.5em}
\caption{\textbf{Comparison of error rate ($\%$) on CIFARC100-C with severity level 5.} We conduct experiments on continual TTA setup. Avg. err means the average error rate (\%) of all 15 corruptions, and Mem. denotes total memory consumption, including model parameter sizes and activations. WRN refers to WideResNet. The implementation details of the baselines are described in \secref{sec:ttaworks}.}
\label{tab:cifar100_all} 
\vspace{-0.4em}
\end{table*}}